\documentclass[journal]{IEEEtran}

\usepackage{xcolor,soul,framed} 

\colorlet{shadecolor}{yellow}
\usepackage[pdftex]{graphicx}
\graphicspath{{../pdf/}{../jpeg/}}
\DeclareGraphicsExtensions{.pdf,.jpeg,.png}

\ifCLASSOPTIONcompsoc
  \usepackage[caption=false,font=normalsize,labelfont=sf,textfont=sf]{subfig}
\else
  \usepackage[caption=false,font=footnotesize]{subfig}
\fi

\usepackage[cmex10]{amsmath}
\usepackage{bm}
\usepackage{array}

\usepackage{tabulary}
\usepackage{mdwmath}
\usepackage{mdwtab}
\usepackage{amsfonts}
\usepackage{amssymb}
\usepackage{amsthm}
\usepackage{eqparbox}
\usepackage{url}
\usepackage{mathtools}
\usepackage{tensor}
\usepackage{multirow}
\usepackage{tikz}
\usetikzlibrary{chains, fit, shapes, decorations.pathreplacing, decorations.pathmorphing, shapes.symbols, calc, trees, positioning, arrows, arrows.meta, shapes.geometric, matrix}
\usepackage{enumerate}
\usepackage{hyperref}
\usepackage{algorithm}
\usepackage{algorithmic}
\usepackage{cite}

\allowdisplaybreaks
\newtheorem{assumption}{Assumption}

\begin{document}
\title{Design and Simulation of Time-energy Optimal Anti-swing Trajectory Planner for Autonomous Tower Cranes}
\author{
Souravik Dutta~%
and~Yiyu Cai~%
\thanks{This work was funded in part by the Energy Research Institute @NTU (ERI@N) of Nanyang Technological University (NTU), Singapore, through the National Research Foundation fellowship. (\textit{Corresponding author: Yiyu Cai.})}
\thanks{S. Dutta is with the School of Mechanical and Aerospace Engineering (MAE) and Energy Research Institute @NTU (ERI@N) - Interdisciplinary Graduate Programme (IGP) of Nanyang Technological University, Singapore 639798 (e-mail: SOURAVIK001@e.ntu.edu.sg).}
\thanks{Y. Cai is with the School of Mechanical and Aerospace Engineering (MAE), and Energy Research Institute @NTU (ERI@N) of Nanyang Technological University, Singapore 639798 (e-mail: MYYCai@ntu.edu.sg).}
}


\maketitle

\begin{abstract}

For autonomous crane lifting, optimal trajectories of the crane are required as reference inputs to the crane controller to facilitate feedforward control. Reducing the unactuated payload motion is a crucial issue for under-actuated tower cranes with spherical pendulum dynamics. The planned trajectory should be optimal in terms of both operating time and energy consumption, to facilitate optimum output spending optimum effort. This article proposes an anti-swing tower crane trajectory planner that can provide time-energy optimal solutions for the Computer-Aided Lift Planning (CALP) system developed at Nanyang Technological University. The CALP system facilitates displacement-optimal collision-free lifting path planning of robotized tower cranes in autonomous construction sites. The current work introduces a trajectory planning module to the system that utilizes the geometric outputs from the path planning module and optimally scales them with time information. Firstly, analyzing the non-linear dynamics of the crane operations, the tower crane is established as differentially flat. Subsequently, the multi-objective trajectory optimization problems for all the crane operations are formulated in the flat output space through consideration of the mechanical and safety constraints. Two multi-objective evolutionary algorithms, namely Non-dominated Sorting Genetic Algorithm (NSGA-II) and Generalized Differential Evolution 3 (GDE3), are extensively compared as potential optimizers. Statistical measures of performance based on the closeness of solutions to the Pareto front, distribution of solutions in the solution space and the runtime of the algorithms are computed to select the optimization engine of the planner. Finally, the crane operation trajectories are obtained via the corresponding planned flat output trajectories. Studies simulating real-world lifting scenarios are conducted to verify the effectiveness and reliability of the proposed module of the lift planning system.

\end{abstract}


\IEEEpeerreviewmaketitle


\section{Introduction}
\label{sec:introduction}

\subsection{Background and Motivation}
\label{subsec:background}

\IEEEPARstart{T}{wo} crucial issues in autonomous crane lift motion planning are path planning and trajectory planning. Path planning generates a collision-free geometric path of the payload, from an initial to a final location, passing through way-points. Trajectory planning takes a given geometric path and scales the profile with time information. Tower cranes are typically under-actuated systems since the unactuated payload swing motion is generated as a result of the independent actuator inputs for the trolley or jib positioning. Consequently, the system dynamics become non-linearly coupled. Trajectory planning is considered a feedforward control method for the crane, where carefully planned trajectories of the trolley/slew operation paths can reduce this swing during lifting. Particular consideration is given to the residual swing of the payload at the end of an operation, even after the actuator force is removed. The residual oscillations of payload swing contribute towards unwanted vibrations of the crane components. In addition, high amplitudes of payload swing during actuated motion can render a planned path infeasible, as certain portions of the path might no longer stay collision-free. To minimize lifting resources, the designed trajectory of a lifting path should also be optimal. The typical optimality criteria for crane lifting trajectories are total operating time and total operating energy. An optimal trajectory with motion-induced swing suppression and residual swing elimination ensures both the safety and productivity of the lifting task.

Lift planning systems focus on the automatic computation of feasible or optimal lift motion plans with minimal user interventions. Crane controllers can utilize the planned lift motions to execute lifting tasks specified by the start and end positions of the building components. Proper environmental representation of the construction site and the building data, including robotic models of cranes, are needed for the computational simulations. The Computer-Aided Lift Planning (CALP) system designed for autonomous cranes at Nanyang Technological University utilizes a combination of Building Information Modeling (BIM), single-level depth map (SDM), online collision detection and massively parallelized genetic algorithm (MSPGA) for generating displacement-optimal collision-free lifting paths in dynamic construction environments \cite{Cai2016,Dutta2020}. Decoupled actuations along the lifting paths are considered to address procedural limitations of heavy-lifting in real-world construction sites. Suitable trajectories are needed for the execution of lifting tasks according to the planned paths obtained from the CALP system. The actuated operations along the paths can be parameterized in terms of the operating time, with polynomial or trigonometric or spline functions. Defining the window of time for each operation brings motion constraints into play, such as limits on position, velocity, and acceleration, for trajectory generation. To ensure avoidance of excessive acceleration of actuators and vibrations of the mechanical structure operating at reasonable speed during heavy lifting, smooth trajectories are required. The solution to the trajectory problem generates necessary reference inputs for the tracking control system of the tower crane so that the desired lifting task can be performed.

Lifting trajectory planning has been investigated in the past extensively for overhead cranes \cite{Sun2012,Hoang2014,Zhang2014,Wu2014,Chen2016,Chen2017,Liu2018,Zhang2021}. The relatively limited research in the literature on tower crane trajectory planning is due to the slewing motion of the jib producing additional inertial and centrifugal forces on the cable-suspended payload. This results in a highly non-linear system with increased coupling between all the degrees of freedom (DOFs), actuated and unactuated. To simultaneously accomplish the complicated objectives of time-specified lifting path realization and payload swing rejection, researchers have applied a number of techniques, as discussed in the next section.

\subsection{Literature Review}
\label{sec:literature}

Most of the recent work looking into the trajectory planning for crane operations have been focused on overhead cranes, which are simpler in their function as there are two actuated degrees of mobility, i.e. the hoisting and the horizontal trolley movement, whereas, in a tower crane, there is the additional slewing operation. Sun et al. \cite{Sun2012} have presented an offline planning method by introducing an anti-swing mechanism into an S-shape reference trajectory for overhead cranes. A rigorous analysis of the coupling behaviour between the trolley motion and the unactuated payload swing is done. The combined trajectory is fine-tuned for accurate trolley positioning using an iterative learning strategy. Another feasible and efficient technique by Hoang et al. \cite{Hoang2014} have utilized a staircase form of trolley acceleration. A scalable number of stairs is computed by calculating the residual oscillation amplitude according to the number of stairs and the transient natural frequency of the cable-suspended payload system. The payload swings in the constant velocity phase are efficiently reduced with the trolley reaching the target position.

Among optimization approaches, Zhang et al. \cite{Zhang2014} have proposed a novel minimum-time trajectory planning solution for overhead cranes, which adopts the quasi-convex optimization technique by means of system discretization and augmentation. The study on overhead crane trajectory planning by Wu and Xia \cite{Wu2014} have aimed at finding an optimal solution of trajectory planning in terms of energy efficiency. Using the optimal control method, an optimal trajectory is obtained with less energy consumption than the compared trajectories, while also satisfying physical and practical constraints such as swing, acceleration and jerk. Chen et al. \cite{Chen2016} have defined a flat output of an overhead crane system based on the differential flatness technique, to deal with the coupling between the payload swing and trolley motion. The trajectory is parameterized by a B-spline curve with unknown parameters while considering the continuity and smoothness requirements. Another work by Chen et al. \cite{Chen2017} has dealt with the double-pendulum model for the overhead crane considering swings of the hook and the payload separately. After linearizing and discretizing the system, careful analysis involving the state constraints presents the trajectory planning problem as a quasi-convex optimization problem, which is solved via a bi-section method. A somewhat similar strategy can be observed in \cite{Liu2018}, where the optimal trajectory of the discretized overhead crane system is deduced via a combination of quintic Bézier curve parameterization and Particle Swarm Optimization (PSO) \cite{Kennedy1995}. The constraints used in the optimization are payload swing angle, trolley velocity, and trolley acceleration. Recently, Zhang et al. \cite{Zhang2021} have generated time-optimal trolley trajectories for a double-pendulum overhead crane system that can avoid obstacles using the hoist operation. The complex non-linear dynamic equations of the double-pendulum crane system are analyzed and a flat output is constructed, which is used to deal with the coupling between the state variables.

The majority of the under-actuated tower crane work in literature addresses payload swing suppression as a closed-loop control problem. Vaughan et al. \cite{Vaughan2010} have reviewed a command generation technique to suppress the oscillatory dynamics of double-pendulum tower cranes, with robustness to frequency changes. However, only the trolley motion is investigated in their work. In \cite{Devesse2013}, a time-optimal velocity control strategy is implemented to achieve fast swing-free tower crane movements by applying Pontryagin's maximum principle on the spherical-pendulum payload equations decomposed as two planar pendulums. The controller is prepared in a feedback form and is employable in real-time. Sun et al. \cite{Sun2016} have presented an adaptive control scheme developed by incorporating some elaborately constructed terms of the system storage function to improve the transient control performance and to reduce unexpected overshoots for the jib and trolley movements. The control scheme is designed without linearizing the tower crane's non-linear dynamics or neglecting partial non-linearities.  Chen et al. \cite{Chen2019} have proposed an adaptive tracking control method, which exhibits satisfactory tracking performance w.r.t. parameter uncertainties and external disturbances. Specifically, exploiting the passivity property, a shaped energy-like function is synthesized as a Lyapunov candidate, based on which, an adaptive tracking controller is developed. A 7\textsuperscript{th}-order polynomial is used to represent the reference trajectory, without optimality consideration. Zhang et al. \cite{Zhang2020} have designed an adaptive integral sliding mode control method with payload sway reduction for 4-DOF tower cranes, where external disturbances, unmodeled uncertainties and parametric uncertainties are present. The designed controller employs an integral sliding mode control strategy to gain robustness and an adaptive control method for performance. Another adaptive tracking controller has been designed by Ouyang et al. \cite{Ouyang2020} for double-pendulum tower cranes. Here, again a reference trajectory is selected based on just the position requirements of the path and kinematic constraints.

The scarce research on tower crane trajectory generation can be categorized based on the inclusion or exclusion of optimality. A new time-(sub)optimal trajectory planning method is proposed by Liu et al. \cite{Liu2019} to generate anti-swing trajectories of 4-DOF tower cranes with state constraints. In particular, three auxiliary signals are constructed, and the reference trajectories for the jib slew and trolley translation can be obtained by the elaborately designed optimal trajectories related to auxiliary signals. A bi-section method is used to solve for the auxiliary trajectory optimization. Ouyang et al. \cite{Ouyang2020a} have designed a composite trajectory planning strategy for tower cranes with a double-pendulum effect. The first part of the proposed trajectory ensures the positioning of the jib and trolley, and the second part secures swing suppression by using a damping component.

\subsection{Limitations of Existing Research}
\label{subsubsec:researchgap}

There are fewer open-loop control methods formed to address the issue of payload swing suppression of tower cranes compared to that of overhead cranes. Most of the proposed closed-loop-control solutions work with only feasible reference trajectories, based on the kinematic constraints of motion while achieving the effective payload swing reduction induced by the crane actuations via feedback control. Some of these approaches require extensive linearization of the payload motions, while others address the coupling non-linear behavior completely. A noticeable lack of boundary constraints for the jerk of the actuated motions is also an issue. Having a jerk-continuous trajectory with zero boundary conditions improves the performance of the system by reducing stress in the actuators and vibrations of the crane components, which in turn decreases the effect of payload swing.

Moreover, there is no solution for multi-objective optimal trajectory planning, as all the previous optimization techniques have been used to achieve only minimum time of operation while constraining the unactuated payload motion during lifting. None of the research in the literature has looked into minimizing the energy spent during lifting operations, which is pertinent in spending optimal effort to gain optimal output. In addition, tracking energy-optimal trajectories is relatively un-challenging, and they also generate less stress on the actuators and the mechanical structure of the crane. Hence, there is a need for a trajectory planning method for autonomous tower crane lifting which can address the non-linearity of the slew operation, minimize the operating time as well as the operating energy, constrain the payload swing during lifting, and eliminate the residual swing at the end of the actuated motion - while satisfying the boundary state constraints. Such a trajectory planner aiding tower cranes in autonomous construction environments can increase the safety and productivity of the construction process, by providing the necessary optimal reference inputs to the crane controller.

\subsection{Contributions of the Proposed Method}
\label{subsec:researchcontributions}

The current research proposes an offline anti-swing trajectory planner for the lift planning system developed at Nanyang Technological University, Singapore \cite{Cai2016,Huang2018}, which is used for lift planning of robotized tower cranes in autonomous construction sites. The existing path planner of the Computer-Aided Lift Planning (CALP) system generates a geometric path consisting of a series of crane operations in C-space. Using the crane configurations at the operation switching instants as inputs, the trajectory planner is designed to list the individual optimal trajectory of each crane operation, in the form of a time-scaled ordered set of position, velocity and acceleration, subjected to limits on the motions, both actuated and unactuated. In this method, the coupling and non-linearity of the state variables are tackled by constructing flat outputs for the trolley-payload and the jib-payload systems. Detailed analysis of the non-linear dynamics of the tower crane is done to prove the differential flatness of the system. The systematic process of trajectory planning for the crane operations is addressed by formulating constrained multi-objective optimization problems with both operating time and energy as the optimality criteria. A multi-objective optimizer is employed to solve the optimization problem to generate optimal lifting trajectories with no residual payload swing, via the flat output space. Finally, the verification and reliability of the developed trajectory planning technique is evaluated by performing simulation studies on lifting operations provided by the lifting path planner of the CALP system.

\section{Robotized Model of the Tower Crane}
\label{sec:cranemodel}

In order to accurately simulate the tower crane behaviour, a robotized crane model has been developed in this section, which captures the most important kinematic and dynamic aspects of the tower crane system.

\subsection{Kinematic Model}
\label{subsec:kinematicmodel}

\begin{figure}[tbp]
     \begin{center}   
        \includegraphics[width=\columnwidth]{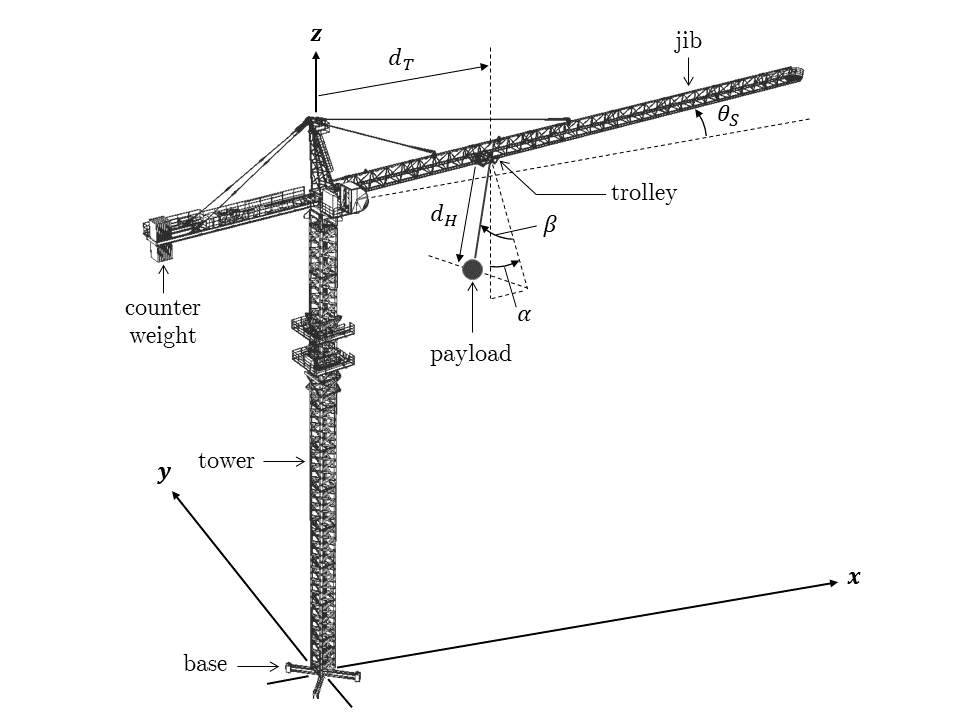}
        \caption{Structure of the robotized tower crane with actuated and unactuated DOFs.}
        \label{fig:craneDOFs}
    \end{center}
\end{figure}

The robotized tower crane is under-actuated with 5 DOFs, as illustrated in Figure \ref{fig:craneDOFs}, where the global (fixed) co-ordinate system $xyz$ is located at the crane base. It consists of joint-connected rigid bodies coupled with a spherical pendulum system. The rigid bodies are the tower, the jib (with the counterweight) and the trolley. The dimensions of the trolley are negligible relative to the other crane components and hence it assumes a lumped mass. There are three actuated motions (crane operations), viz. slew angle $\theta_S$ about the $z$-axis on the $xy$ plane, trolley distance $d_T$ along the jib of the crane, and hoist length $d_H$ of the cable-payload unit. $\theta_S$ is taken as positive for counter-clockwise rotation of the jib about the $z$ axis when seen from the top (along the $-z$ direction). The unactuated motions are the payload swing angles determined by the crane motion dynamics during lifting operations. $\alpha$ denotes the first swing angle of the payload formed between the $z$-axis and the projection of the hoist-cable on the plane created by the $z$-axis and the jib. The second swing angle of the payload is defined by $\beta$, formed between the hoist-cable and its projection on the plane created by the $z$-axis and the jib. Evidently, $\beta$ is on the plane perpendicular to the plane $\alpha$ belongs to.

The following assumptions regarding tower crane lifting are made to prepare the robotized model of the crane:

\begin{assumption}
    The hoist-cable is massless and does not twist or stretch under load. The hook, having insignificant mass compared to the payload is considered as a part of the cable itself. The payload is also modelled as a lumped mass. The payload and the hoist-cable together constitute the spherical pendulum system. The payload acts as the pendulum which can swing along two unactuated DOFs.
\end{assumption}
    
\begin{assumption}
    Throughout the lifting process, the payload oscillations never exceed the horizontal level of the jib-trolley section \cite{Sun2012a,Sun2013a,Huang2013}. Therefore:
    
    \begin{equation}
        \alpha,\beta \in \left(-\frac{\pi}{2},\frac{\pi}{2}\right)
        \label{eq:swinglimits}
    \end{equation}
\end{assumption}

\begin{assumption}
     The operations of the tower crane are decoupled, i.e. at a time, there is only one specified actuation along the corresponding DOF. No two operations are performed simultaneously \cite{Dutta2020}.
\end{assumption} 

\begin{assumption}
    Any frictional effect on the actuated motions as well as air resistance on the payload is negligible. No strong wind is present during the execution of the lifting task.
\end{assumption}

\subsection{Dynamic Model}
\label{subsec:dynamicmodel}

The tower crane is defined as a system of coupled rigid bodies. The jib of the tower crane is modelled as a rotating beam with a mass moment of inertia $I_J$ about the $z$ axis. The trolley is modelled as a point mass $m_T$, and its motion is constrained to move along the jib. The massless and inextensible hoist-cable is attached to the trolley at one end and to the payload of point mass $m_L$ at the other end. Step-less speed electric motors impart motion along the three actuated DOFs through the following torques and forces: torque $\tau_S$ generating from the slew actuator acting on the jib, force $F_T$ generating from the trolley actuator acting on the trolley and force $F_H$ generating from the hoist actuator acting on the payload.

The equations of motion of the tower crane for each operation: slewing, trolley transportation and hoisting, can be derived using the Lagrangian modelling technique \cite{Goldstein2002}. A Lagrangian function $\mathcal{L}$ is defined as:

\begin{equation} 
     \mathcal{L} = \mathcal{T} - \mathcal{V}
     \label{eqn:lagrangian}
\end{equation}

Here, $\mathcal{T}$ is the total kinetic energy and $\mathcal{V}$ is the total potential energy of a system. The Euler-Lagrange equations for a system with $m$ active DOFs (generalized coordinates) are:

\begin{equation} 
    \frac{d}{dt}\frac{\partial \mathcal{L}}{\partial \Dot{q_i}} - \frac{\partial \mathcal{L}}{\partial {q_i}} = Q_i; \qquad i=1,2,3,...,m
    \label{eqn:genmotionequation}
\end{equation}

In this formula, $q_i$ are the generalized coordinates. The term $Q_i$ represents the generalized forces and is the sum of the external and non-conservative torques and forces acting along the respective generalized coordinates.

Since the tower crane operations are decoupled, different sets of DOFs are active during individual operations, leading to different expressions of kinetic and potential energies. Individual Lagrangian functions and subsequent equations of motion for each operation can be obtained separately. Since the primary attention of the current work is to plan efficient trajectories to suppress the payload swing, exploration of the dynamics of the trolley and slew motions are emphasized below.

\subsubsection{Trolley operation dynamics}
\label{subsubsec:trolleydynamics}

For pure trolley motion, only the in-plane swing angle $\alpha$ is excited due to the actuated movement, exhibiting a planar pendulum response. Let $D_H$ be the fixed length of the hoist-cable during the trolley transportation. The resulting Euler-Lagrange equations for trolley operation can then be written as:

\begin{equation}
    (m_T+m_L)\ddot{d_T} + m_L\ddot{\alpha}D_H\cos{\alpha} - m_L\dot{\alpha}^2D_H\sin{\alpha} = F_T
    \label{eq:trolleydT}
\end{equation}

\begin{equation}
    D_H\ddot{\alpha} + \ddot{d_T}\cos{\alpha} + g\sin{\alpha} = 0
    \label{eq:trolleyalpha}
\end{equation}

Here, $d_T$, $\alpha$ and $F_T$ are all functions of time $t$, and $g$ denotes the gravitational acceleration. Given, that the payload swing has to be kept as low as possible, it can be assumed that the small swing angle approximation ($\alpha < 5^\circ$) is satisfied during the operation \cite{Sun2013}. This implies $\sin{\alpha} \approx \alpha$ and $\cos{\alpha} \approx 1$. Utilizing this, the payload swing equation during trolley translation can be simplified as:

\begin{equation}
    D_H\ddot{\alpha} + \ddot{d_T} + g\alpha = 0
    \label{eq:trolleyalphasimp}
\end{equation}

Eq. (\ref{eq:trolleyalphasimp}) clearly indicates the coupling relationship between the trolley acceleration and the payload swing angle. For the simple pendulum system, the trolley acceleration can be considered as the input and the payload swing angle as the output.

\subsubsection{Slew operation dynamics}
\label{subsubsec:slewdynamics}

For pure slew motion, the additional inertial and centrifugal forces of the slewing action of the crane jib generate both radial component $\alpha$ and the tangential component $\beta$ of the payload swing. As a result, the payload demonstrates a spherical pendulum motion. Let $D_T$ and $D_H$ be the fixed distance of the trolley along the jib and the fixed hoisting height while slewing, respectively. Therefore, the equations of motion for the slew operation are expressed according to the Lagrangian method:

\begin{equation}
    \begin{aligned}
        & (I_J+m_TD_T^2+m_LD_T^2\sin^2{\beta})\ddot{\theta_S} \\
        & + 2m_LD_TD_H\dot{\theta_S}\dot{\alpha}\cos{\beta}\sin^2{\beta}\cos{\alpha} \\
        & - m_LD_TD_H\dot{\theta_S}^2\sin{\beta} - 2m_LD_TD_H\dot{\theta_S}\dot{\beta}\sin{\beta}\sin{\alpha} \\
        & - m_LD_TD_H\dot{\alpha}^2\sin{\beta}\cos^2{\beta} - m_LD_TD_H\dot{\beta}^2\sin{\beta} \\
        & - m_LD_T^2\dot{\alpha}^2\sin{\beta}\cos{\beta}\sin{\alpha} \\
        & + m_LD_TD_H\dot{\alpha}^2\sin{\beta}\cos^2{\beta}\cos^2{\alpha} \\
        & -m_LgD_T\sin{\beta}\cos{\beta}\cos{\alpha} = \tau_S
    \end{aligned}
    \label{eq:slewthetaS}
\end{equation}

\begin{equation}
    \begin{aligned}
        & D_H\ddot{\alpha}\cos{\beta} - D_H\ddot{\theta_S}\sin{\beta}\cos{\alpha} - 2D_H\dot{\alpha}\dot{\beta}\sin{\beta} \\
        & - 2D_H\dot{\beta}\dot{\theta_S}\cos{\beta}\cos{\alpha} - D_H\dot{\theta_S}^2\cos{\beta}\sin{\alpha}\cos{\alpha} \\
        & - D_T\dot{\theta_S}^2\cos{\alpha} + g\sin{\alpha} = 0
    \end{aligned}
    \label{eq:slewalpha}
\end{equation}

\begin{equation}
    \begin{aligned}
        & D_H\ddot{\beta} + (D_T\cos{\beta}+D_H\sin{\alpha})\ddot{\theta_S} + D_H\dot{\alpha}^2\sin{\beta}\cos{\beta} \\
        & - D_H\dot{\theta_S}^2\sin{\beta}\cos{\beta}\cos^2{\alpha} + 2D_H\dot{\theta_S}\dot{\alpha}\cos^2{\beta}\cos{\alpha} \\
        & + D_T\dot{\theta_S}^2\sin{\beta}\sin{\alpha} + g\sin{\beta}\cos{\alpha} = 0
    \end{aligned}
    \label{eq:slewbeta}
\end{equation}

Here, $\theta_S$, $\alpha$, $\beta$ and $\tau_S$ are all functions of time $t$. Now, applying the small swing angle approximation to both the swing angles, i.e. $\sin{\alpha} \approx \alpha$, $\cos{\alpha} \approx 1$, $\dot{\alpha}^2 \approx 0$, $\alpha\beta \approx 0$ and $\dot{\alpha}\dot{\beta} \approx 0$, the equations for the radial and tangential swing angles in the course of the slew process becomes:

\begin{equation}
    D_H\ddot{\alpha} - D_H\ddot{\theta_S}\beta - 2D_H\dot{\beta}\dot{\theta_S} - D_H\dot{\theta_S}^2\alpha - D_T\dot{\theta_S}^2 + g\alpha = 0
    \label{eq:slewalphasimp}
\end{equation}

\begin{equation}
        D_H\ddot{\beta} + (D_T+D_H\alpha)\ddot{\theta_S} - D_H\dot{\theta_S}^2\beta + 2D_H\dot{\theta_S}\dot{\alpha} + g\beta = 0
    \label{eq:slewbetasimp}
\end{equation}

As can be observed, even after employing the small angle approximation, the motions involve terms denoting centrifugal force and Coriolis force components. As a result, the payload swing equations are still non-linear with coupling between the state variables.

\section{Multi-objective Trajectory Optimization Problem}
\label{sec:trajectoryoptimizationproblem}

To deduce optimal trajectories for the actuated DOFs of the tower crane with minimized time and energy of operation, certain constraints need to be considered from the perspective of the mechanical limits of the crane and the motion safety. These constraints need to be satisfied first for the generated trajectory to be a feasible one.

\begin{enumerate}
    \item At the beginning of an operation, the actuators are at rest. Certain kinematic properties of the DOFs (actuated and unactuated) should be zero at this position. Without loss of generalization, this time instant is set as $t=0$. The velocities, accelerations and jerks of the actuated DOFs (hoist, trolley, slew) should be zero, while the position and velocity of the swing angles should be zero too. The initial hoist length, trolley position and slew angle are determined by the corresponding operation-switching configuration of the crane (in the case of decoupled operations, the end of one operation denotes the start of another).
    
    \item Similarly, at the end of an operation, all the aforementioned kinematic properties of the DOFs should become zero. Particular importance is given to the payload swing angles and their velocities, which is one of the open-loop control objectives of the trajectory planning method. Only by restricting both the swing angles and their velocities to zero value at the end of the actuated process, the elimination of residual oscillations can be achieved. As can be noted due to the decoupled lifting path, zero conditions are in effect at each operation switching point.
    
    \item Throughout the course of an operation, due to mechanical limitations of the crane components and actuators, velocities and accelerations of the hoist-cable, trolley and jib are bounded by respective maximum attainable values. These values are usually obtained from the crane manufacturer's specifications. Moreover, the crucial trolley/jib motion-induced payload swing should also be within an allowable limit. This limit can be determined by the permissible deflection of the payload to stay within the collision-free zone of the planned lifting path.
\end{enumerate}

Considering the aforementioned constraints, the multi-objective trajectory optimization problem (MOTOP) for each operation can be formulated.

\subsection{Hoist Trajectory Optimization Problem}
\label{subsec:hoistoptimizationproblem}

Considering $t_H$ as the total hoisting (up or down) time, the position, velocity and acceleration of the hoist action at the initial and final stages of the operation are given by the following constraints:

\begin{equation}
    d_H(0) = d_{Hi}, \; \dot{d_H}(0) = 0, \; \ddot{d_H}(0) = 0, \; d_H^{(3)}(0) = 0
    \label{eq:hoisteqconstinitial}
\end{equation}

\begin{equation}
    d_H(t_H) = d_{Hf}, \; \dot{d_H}(t_H) = 0, \; \ddot{d_H}(t_H) = 0, \; d_H^{(3)}(t_H) = 0
    \label{eq:hoisteqconstfinal}
\end{equation}

In these constraint equations, $d_{Hi}$ and $d_{Hf}$ are the initial and final hoisting distance of the payload from the jib, respectively.

In the absence of external disturbances such as strong wind, there is no swing of the payload during the operation. The payload is hoisted vertically and its movement is along the length of the hoist-cable. As the payload is hoisted up or down, its velocity and acceleration are to be kept inside the permissible range.

\begin{equation}
        \max_{t\in[0,t_H]} \, |\dot{d_H}(t)| \leq \overline{\dot{d_H}}, \; \max_{t\in[0,t_H]} \, |\ddot{d_H}(t)| \leq \overline{\ddot{d_H}}
        \label{eq:hoistineqconst}
\end{equation}

In the aforementioned constraint equation, $\overline{\dot{d_H}}$ and $\overline{\ddot{d_H}}$ are the limits on the velocity and acceleration of hoisting action, respectively.

In order to optimize the hoisting time, $t_H$ needs to be minimized. The energy expense of the operation can be expressed in terms of the total actuator effort, which is the total squared hoisting actuator force $\int_0^{t_H} \left|F_H(t)\right|^2dt$. Since the actuator force is directly affecting the hoisting acceleration, the integral of the squared acceleration $\int_0^{t_H} \left|\ddot{d_H}(t)\right|^2dt$ can also pose as a measure of the consumed energy. It is normalized with respect to the maximum allowed acceleration of the actuator. Therefore, the hoisting trajectory is to be solved for the MOTOP defined below.

\begin{equation}
    \begin{aligned}
        \min _{t_H>0}\; & \left[ t_H, \; \displaystyle\int_0^{t_H} \frac{\left| \ddot{d_H}(t) \right|^2}{\overline{\ddot{d_H}}^2} \; dt \right]^\textrm{T} \\
        \textrm{s.t.} \; & \max_{t\in[0,t_H]} \, \left| \dot{d_H}(t) \right| \leq \overline{\dot{d_H}} \\
        & \max_{t\in[0,t_H]} \, \left| \ddot{d_H}(t) \right| \leq \overline{\ddot{d_H}} \\
        & d_H(0) = d_{Hi}, \; \dot{d_H}(0) = \ddot{d_H}(0) = d_H^{(3)}(0) = 0 \\
        & d_H(t_H) = d_{Hf}, \; \dot{d_H}(t_H) = \ddot{d_H}(t_H) = d_H^{(3)}(t_H) = 0
    \end{aligned}
    \label{eq:hoistoptprob}
\end{equation}

Analyzing the optimization problem, the 8 equality constraints can be used to parameterize the hoist trajectory $d_H(t)$ in terms of $t_H$ via a 7\textsuperscript{th} degree polynomial. B{\'e}zier curves provide a robust computational approach to solve for the coefficients of such high degree polynomials \cite{LuigiBiagiotti2008}. It is to be noted, that a 7\textsuperscript{th} degree polynomial trajectory has jerk continuity throughout the operation time, inclusive of the endpoints.

\subsubsection{Hoist trajectory parameterization}
\label{subsubsec:hoistparamterization}

The hoist length can be defined in the normalized B{\'e}zier form as:

\begin{equation}
    d_H(\eta_H) = \displaystyle\sum_{r=0}^{7} \tensor[^r]{b}{_H}B_r^7(\eta_H); \qquad 0 \leq \eta_H \leq 1
\end{equation}
\label{eq:hoisttrajectoryBezier}

Here, $\eta_H=\frac{t}{t_H}$ denotes the normalized time. The coefficients $\tensor[^r]{b}{_H}$ are the control points of the B{\'e}zier curve representing the hoist trajectory, and the basis functions $B_r^7(\eta_H)$ are 7\textsuperscript{th} degree Bernstein Polynomials defined as $B_r^7(\eta_H)=\frac{7!}{r!(7-r)!} \eta_H^r(1-\eta_H)^{7-r}$.

According to the equality constraints (boundary conditions) represented by Eqs. (\ref{eq:hoisteqconstinitial} - \ref{eq:hoisteqconstfinal}), the control points of the normalized Bézier form of the hoist trajectories are obtained.

\begin{equation}
    \begin{aligned}
        & \tensor[^0]{b}{_H} = d_{Hi}, \; \tensor[^1]{b}{_H} = d_{Hi}, \; \tensor[^2]{b}{_H} = d_{Hi}, \; \tensor[^3]{b}{_H} = d_{Hi},\\
        & \tensor[^4]{b}{_H} = d_{Hf}, \; \tensor[^5]{b}{_H} = d_{Hf}, \; \tensor[^6]{b}{_H} = d_{Hf}, \; \tensor[^7]{b}{_H} = d_{Hf}
    \end{aligned}
    \label{eq:hoisttrajectorycps}
\end{equation}

Substituting the values of $\tensor[^r]{b}{_H}$ for $r=0,...,7$, the hoist trajectory $d_H(t)$ is derived as a polynomial function of time in terms of $t_H$.

\begin{equation}
    d_H(t) = \displaystyle\sum_{r=0}^{7} \tensor[^r]{c}{_H}\left(\frac{t}{t_H}\right)^r; \qquad 0 \leq t \leq t_H
    \label{eqe:hoisttrajectorypoly}
\end{equation}

Expressing the total traversed hoisting height as $\Delta d_H = d_{Hf}-d_{Hi}$, the values of the coefficients $\tensor[^r]{c}{_H}$ are determined as the following:

\begin{equation}
    \begin{aligned}
        & \tensor[^0]{c}{_H} = d_{Hi}, \; \tensor[^1]{c}{_H} = 0, \; \tensor[^2]{c}{_H} = 0, \; \tensor[^3]{c}{_H} = 0, \; \tensor[^4]{c}{_H} = 35\Delta d_H, \\
        & \tensor[^5]{c}{_H} = -84\Delta d_H, \; \tensor[^6]{c}{_H} = 70\Delta d_H, \; \tensor[^7]{c}{_H} = -20\Delta d_H
    \end{aligned}
    \label{eq:hoisttrajectorycoeff}
\end{equation}  

Subsequently, the hoist velocity trajectory $\dot{d_H}(t)$ and the hoist acceleration trajectory $\ddot{d_H}(t)$ are also derived in their parameterized version, via successive differentiation of $d_H(t)$. Hence, the hoist trajectory optimization problem defined in Eq. (\ref{eq:hoistoptprob}) essentially reduces to a MOTOP with only inequality constraints:

\begin{equation}
    \begin{aligned}
        \min _{t_H>0}\; & \left[ f_1(t_H), \; f_2(t_H) \right]^\textrm{T}\\
        \textrm{s.t.} \; & \textrm{Eq. (\ref{eq:hoistineqconst})}
    \end{aligned}
    \label{eq:hoistMOTOP}
\end{equation}

In the above equations, $d_H(t)$ is computed according to Eq. (\ref{eqe:hoisttrajectorypoly}), whose coefficients are calculated as Eq. (\ref{eq:hoisttrajectorycoeff}). The solution to the MOTOP for hoist operation is the corresponding value of $t_H$ for the optimal hoisting time and energy.

\subsection{Trolley Trajectory Optimization Problem}
\label{sec:trolleyoptimizationproblem}

Let $t_T$ be the total trolley transportation time. The position, velocity and acceleration of the trolley at the initial and final frames of the motion have to abide by the following values:

\begin{equation}
    d_T(0) = d_{Ti}, \; \dot{d_T}(0) = 0, \; \ddot{d_T}(0) = 0, \; d_T^{(3)}(0) = 0
    \label{eq:trolleyeqconstinitial}
\end{equation}

\begin{equation}
    d_T(t_T) = d_{Tf}, \; \dot{d_T}(t_T) = 0, \; \ddot{d_T}(t_T) = 0, \; d_T^{(3)}(t_T) = 0
    \label{eq:trolleyeqconstfinal}
\end{equation}  

Here, $d_{Ti}$ and $d_{Tf}$ are the initial and final positions of the trolley along the jib, respectively.

At the same time, the planar swing angle of the payload should also be restricted at the boundaries of the motion as:

\begin{equation}
    \alpha(0) = 0, \; \dot{\alpha}(0) = 0
    \label{eq:trolleyswingeqconstinitial}
\end{equation}

\begin{equation}
    \alpha(t_T) = 0, \; \dot{\alpha}(t_T) = 0
    \label{eq:trolleyswingeqconstfinal}
\end{equation}  

As the trolley translates, its velocity and acceleration are to be kept within a specified range. Similarly, the transient amplitude of the swing angle should be controlled based on the maximum allowed deflection of the payload.

\begin{equation}
    \max_{t\in[0,t_T]} \, \left| \dot{d_T}(t) \right| \leq \overline{\dot{d_T}}, \; \max_{t\in[0,t_T]} \, \left| \ddot{d_T}(t) \right| \leq \overline{\ddot{d_T}}
    \label{eq:trolleyineqconst}
\end{equation}

\begin{equation}
    \max_{t\in[0,t_T] \, }\left| \alpha(t) \right| \leq \overline{\alpha}
    \label{eq:trolleyswingineqconst}
\end{equation}  

In the aforementioned equations, $\overline{\dot{d_T}}$, $\overline{\ddot{d_T}}$ and $\overline{\alpha}$ are the maximum allowed values of the trolley velocity, the trolley acceleration, and the payload swing angle, respectively.

Then, the MOTOP for the trolley operation can be constructed as:

\begin{equation}
    \begin{aligned}
        \min_{t_T>0} \; & \left[ t_T, \; \displaystyle\int_0^{t_T} \frac{\left| \ddot{d_T}(t) \right|^2}{\overline{\ddot{d_T}}^2} \; dt \right]^\textrm{T} \\
        \textrm{s.t.} \; & \max_{t\in[0,t_T]} \, \left| \dot{d_T}(t) \right| \leq \overline{\dot{d_T}} \\
        & \max_{t\in[0,t_T]} \, \left| \ddot{d_T}(t) \right| \leq \overline{\ddot{d_T}} \\
        & \max_{t\in[0,t_T]} \, \left| \alpha(t) \right| \leq \overline{\alpha} \\
        & d_T(0) = d_{Ti}, \; \dot{d_T}(0) = \ddot{d_T}(0) = d_T^{(3)}(0) = 0 \\
        & d_T(t_T) = d_{Tf}, \; \dot{d_T}(t_T) = \ddot{d_T}(t_T) = d_T^{(3)}(t_T) = 0 \\
        & \alpha(0) = 0, \; \dot{\alpha}(0) = 0 \\
        & \alpha(t_T) = 0, \; \dot{\alpha}(t_T) = 0
    \end{aligned}
    \label{eq:trolleyoptprob}
\end{equation}  

After careful analysis of the trolley-payload dynamics represented by Eq. (\ref{eq:trolleyalphasimp}) and the optimization problem constructed above, parameterization of the trolley trajectory $d_T(t)$ is deemed insufficient to formulate a parameterized MOTOP in terms of $t_T$. As the swing angle trajectory $\alpha(t)$ cannot be calculated directly via the parameterized trolley trajectory (or its time derivatives) because of the nonlinear coupling behavior exhibited by the payload dynamics, the inequality and inequality constraints of the MOTOP involving the payload swing introduces additional difficulty. However, this can be solved by defining a flat output of the system formed by the trolley and the simple pendulum.

\subsubsection{Differential flatness of trolley-payload system}
\label{subsubsec:flatoutputtrolley}

According to Fliess et al. \cite{Fliess1992}, a system is differentially flat if there exists an output or there exist a number of outputs, such that the state and input variables can be parameterized by the output(s) and a finite number of its(their) time derivatives. If the differential flatness of the trolley-pendulum system can be proven, the trajectory planning can be done in the flat output space, as all the DOFs (and hence their time derivatives) are represented by the flat output(s) and its(their) time derivatives.

\begin{figure}[tbp]
     \begin{center}   
        \includegraphics[width=\columnwidth]{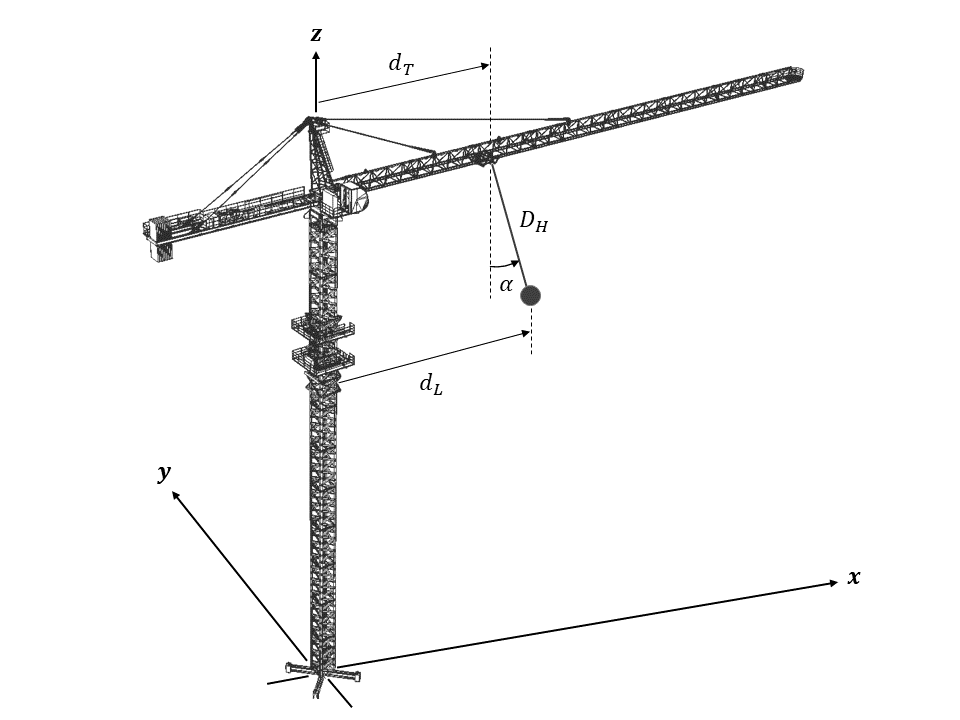}
        \caption{Auxiliary DOF for the trolley-payload system during trolley operation.}
        \label{fig:trolleyauxDOF}
    \end{center}
\end{figure}

To this extent, the distance of the payload along the jib (on the plane created by the jib and the $z$-axis) is considered as an auxiliary DOF of the system. Figure \ref{fig:trolleyauxDOF} shows this DOF as $d_L$. This DOF is an output of the system due to the trolley acceleration.

By geometric analysis of the trolley and payload movements:

\begin{equation}
    d_L = d_T + D_H\sin{\alpha}
    \label{eq:trolleyauxDOF}
\end{equation}  

Approximating the swing angle as a small angle ($\sin{\alpha} \approx \alpha$):

\begin{equation}
    d_L = d_T + D_H\alpha
    \label{eq:trolleyauxDOFsimp}
\end{equation}  

Differentiating twice with respect to (w.r.t.) time:

\begin{equation}
    \ddot{d_L} = \ddot{d_T} + D_H\ddot{\alpha}
    \label{eq:trolleyauxDOFacc}
\end{equation}  

Now, combining Eqs. (\ref{eq:trolleyalphasimp}) and (\ref{eq:trolleyauxDOFacc}), it is deduced that:

\begin{equation}
    \alpha = -\frac{\ddot{d_L}}{g} 
    \label{eq:trolleyalphafp}
\end{equation}  

Substituting $\alpha$ in Eq. (\ref{eq:trolleyauxDOFsimp}) by Eq. (\ref{eq:trolleyalphafp}), it is shown that:

\begin{equation}
    d_T = d_L + \frac{D_H}{g}\ddot{d_L} 
    \label{eq:trolleydTfp}
\end{equation}  

As evident from the two equations above, the actuated DOF $d_T$, and the unactuated DOF $\alpha$ (and their velocities and accelerations via differentiation w.r.t. time), can be represented by the auxiliary DOF $d_L$, and some of its time derivatives. Hence, it can be concluded that $d_L$ is a flat output of the trolley-payload system.

Differentiating Eq. (\ref{eq:trolleydTfp}) twice, and Eq. (\ref{eq:trolleyalphafp}) once, respectively, w.r.t. time, the other state variables to be constrained during the trolley motion are represented as follows:

\begin{equation}
    \dot{d_T} = \dot{d_L} + \frac{D_H}{g}d_L^{(3)} 
    \label{eq:trolleydTvelfp}
\end{equation}

\begin{equation}
    \ddot{d_T} = \ddot{d_L} + \frac{D_H}{g}d_L^{(4)} 
    \label{eq:trolleydTaccfp}
\end{equation}

\begin{equation}
    \dot{\alpha} = -\frac{d_L^{(3)}}{g} 
    \label{eq:trolleyalphavelfp}
\end{equation}  

\subsubsection{Trolley trajectory planning in flat output space}
\label{subsubsec:trolleytrajectoryflatoutput}

With the differential flatness for the tower crane during trolley motion established, the flat output can be manipulated to plan the trolley trajectory by planning the flat output trajectory. So, the trolley MOTOP presented through Eq. (\ref{eq:trolleyoptprob}) is formulated in terms of $d_L$ and its time derivatives.

From the equality constraints given by Eqs. (\ref{eq:trolleyeqconstinitial} - \ref{eq:trolleyswingeqconstfinal}), and using the expressions of the state variables in Eqs. (\ref{eq:trolleyalphafp} - \ref{eq:trolleyalphavelfp}), the following straightforward calculations are obtained:

\begin{equation}
    \begin{aligned}
        & d_L(0) = d_{Ti}, \; \dot{d_L}(0) = 0, \; \ddot{d_L}(0) = 0, \\
        & d_L^{(3)}(0) = 0, \; d_L^{(4)}(0) = 0, \; d_L^{(5)}(0) = 0
    \end{aligned}
    \label{eq:trolleyfpeqconstinitial}
\end{equation}

\begin{equation}
    \begin{aligned}
        & d_L(t_T) = d_{Tf}, \; \dot{d_L}(t_T) = 0, \; \ddot{d_L}(t_T) = 0, \\ 
        & d_L^{(3)}(t_T) = 0, \; d_L^{(4)}(t_T) = 0, \; d_L^{(5)}(t_T) = 0
    \end{aligned}
    \label{eq:trolleyfpeqconstfinal}
\end{equation}  

Substituting the flat output expressions of the state variables in the inequality constraints for the trolley operation formed in Eqs. (\ref{eq:trolleyineqconst}-\ref{eq:trolleyswingineqconst}):

\begin{equation}
\begin{aligned}    
        & \max_{t\in[0,t_T]} \, \left| \dot{d_L}(t) + \frac{D_H}{g}d_L^{(3)}(t) \right| \leq \overline{\dot{d_T}}, \\
        & \max_{t\in[0,t_T]} \, \left| \ddot{d_L}(t) + \frac{D_H}{g}d_L^{(4)}(t) \right| \leq \overline{\ddot{d_T}}
    \end{aligned}
    \label{eq:trolleyfpineqcontprelim}
\end{equation}

\begin{equation}
    \max_{t\in[0,t_T]} \, \left| \frac{\ddot{d_L}(t)}{g} \right| \leq \overline{\alpha}
    \label{eq:trolleyfpswingineqcontprelim}
\end{equation}  

According to the subadditivity property of absolute value inequality:

\begin{equation}
        \max_{t\in[0,t_T]} \, \left| \ddot{d_L}(t) + \frac{D_H}{g}d_L^{(4)}(t) \right| \leq g\overline{\alpha} + \max_{t\in[0,t_T]} \, \left| \frac{D_H}{g}d_L^{(4)}(t) \right|
\end{equation}  

So, Eq. (\ref{eq:trolleyfpineqcontprelim} - \ref{eq:trolleyfpswingineqcontprelim}) can be written as:

\begin{equation}
\begin{aligned}    
        & \max_{t\in[0,t_T]} \, \left| \dot{d_L}(t) + \frac{D_H}{g}d_L^{(3)}(t) \right| \leq \overline{\dot{d_T}}, \\
        & \max_{t\in[0,t_T]} \, \left| d_L^{(4)}(t) \right| \leq \frac{g}{D_H}(\overline{\ddot{d_T}} - g\overline{\alpha})
    \end{aligned}
    \label{eq:trolleyfpineqcont}
\end{equation}

\begin{equation}
    \max_{t\in[0,t_T]} \, \left| \ddot{d_L}(t) \right| \leq g\overline{\alpha}
    \label{eq:trolleyfpswingineqcont}
\end{equation}  

Considering the 12 boundary conditions of the flat output given by Eqs. (\ref{eq:trolleyfpeqconstinitial} - \ref{eq:trolleyfpeqconstfinal}), an 11\textsuperscript{th} degree B\'{e}zier curve can be implemented to parameterize $d_L(t)$. Following the procedure of parameterization explained in Section \ref{subsubsec:hoistparamterization} the flat output for the trolley-payload system is expressed in terms of $t_T$ as:

\begin{equation}
    d_L(t) = \displaystyle\sum_{r=0}^{11} \tensor[^r]{c}{_L}\left(\frac{t}{t_T}\right)^r; \qquad 0 \leq t \leq t_T
    \label{eqe:trolleyfptrajectorypoly}
\end{equation}  

Expressing the total change of the payload position along the direction of trolley movement as $\Delta d_L = d_{Tf}-d_{Ti}$, the values of the coefficients $\tensor[^r]{c}{_L}$ are determined as the following:

\begin{equation}
    \begin{aligned}
        & \tensor[^0]{c}{_L} = d_{Ti}, \; \tensor[^1]{c}{_L} = 0, \; \tensor[^2]{c}{_L} = 0, \; \tensor[^3]{c}{_L} = 0, \; \tensor[^4]{c}{_L} = 0, \; \tensor[^5]{c}{_L} = 0, \\
        & \tensor[^6]{c}{_L} = 462\Delta d_L, \; \tensor[^7]{c}{_L} = -1980\Delta d_L, \; \tensor[^8]{c}{_L} = 3465\Delta d_L, \\
        & \tensor[^9]{c}{_L} = -3080\Delta d_L, \; \tensor[^{10}]{c}{_L} = 1386\Delta d_L, \; \tensor[^{11}]{c}{_L} = -252\Delta d_L
    \end{aligned}
    \label{eq:trolleyfptrajectorycoeff}
\end{equation}  

Subsequently, the trajectories of the time derivatives of the flat output are also derived in their parameterized versions in terms of $t_T$, via successive differentiation of $d_L(t)$. Consequently, the trolley trajectory optimization problem defined in Eq. (\ref{eq:trolleyoptprob}) is converted to a MOTOP of the flat output, with only inequality constraints:

\begin{equation}
    \begin{aligned}
        \min _{t_T>0}\; & \left[ f_1(t_T), \; f_2(t_T) \right]^\textrm{T} \\
        \textrm{s.t.} \; & \textrm{Eqs. (\ref{eq:trolleyfpineqcont} - \ref{eq:trolleyfpswingineqcont})}
    \end{aligned}
    \label{eq:trolleyMOTOP}
\end{equation}

In Eq. (\ref{eq:trolleyMOTOP}), $d_L(t)$ is given by Eq. (\ref{eqe:trolleyfptrajectorypoly}), whose coefficients are calculated as Eq. (\ref{eq:trolleyfptrajectorycoeff}). The solution to the MOTOP for the trolley operation is the corresponding value of $t_T$ for the optimal hoisting time and energy. Once $t_T$ is obtained, the optimal trolley trajectories can be computed using Eqs. (\ref{eq:trolleydTfp} - \ref{eq:trolleydTaccfp}).

\subsection{Slew Trajectory Optimization Problem}
\label{sec:slewoptimizationproblem}

Considering $t_S$ to be the total time for the slewing operation, the angular position, velocity and acceleration of the crane jib at the initial and final frames of the motion are to be constrained within the following values:

\begin{equation}
        \theta_S(0) = \theta_{Si}, \; \dot{\theta_S}(0) = 0, \; \ddot{\theta_S}(0) = 0, \; \theta_S^{(3)}(0) = 0
        \label{eq:sleweqconstinitial}
\end{equation}

\begin{equation}
        \theta_S(t_S) = \theta_{Sf}, \; \dot{\theta_S}(t_S) = 0, \; \ddot{\theta_S}(t_S) = 0, \; \theta_S^{(3)}(t_S) = 0
        \label{eq:sleweqconstfinal}
\end{equation}  

Here, $\theta_{Si}$ and $\theta_{Sf}$ are the initial and final angular positions of the crane jib, respectively.

Also, the radial and tangential swing angles of the spherical pendulum system formed by the hoist-cable and the payload should also be specified at the beginning and end of the slew motion as:

\begin{equation}
        \alpha(0) = \beta(0) = 0, \; \dot{\alpha}(0) = \dot{\beta}(0) = 0
        \label{eq:slewswingeqconstinitial}
\end{equation}

\begin{equation}
        \alpha(t_S) = \beta(t_S) = 0, \; \dot{\alpha}(t_S) = \dot{\beta}(t_S) = 0
        \label{eq:slewswingeqconstfinal}
\end{equation}  

As the jib rotates, its velocity and acceleration are to be confined within their specific limits. At the same time, the transient amplitudes of the swing angles should be constrained based on the limit of permissible diversion of the payload.

\begin{equation}
        \max_{t\in[0,t_S]} \, \left| \dot{\theta_S}(t) \right| \leq \overline{\dot{\theta_S}}, \; \max_{t\in[0,t_S]} \, \left| \ddot{\theta_S}(t) \right| \leq \overline{\ddot{\theta_S}}
    \label{eq:slewineqconst}
\end{equation}

\begin{equation}
        \max_{t\in[0,t_S]} \, \left| \alpha(t) \right| \leq \overline{\alpha}, \; \max_{t\in[0,t_S]} \, \left| \beta(t) \right| \leq \overline{\beta}
    \label{eq:slewswingineqconst}
\end{equation}  

In these equations, $\overline{\dot{\theta_S}}$, $\overline{\ddot{\theta_S}}$, $\overline{\alpha}$ and $\overline{\beta}$ are the maximum attainable values of the slew velocity, the slew acceleration, the radial payload swing angle, and the tangential payload swing angle, respectively.

So, the MOTOP for the slew trajectory can be constructed as:

\begin{equation}
    \begin{aligned}
        \min_{t_S>0} \; & \left[ t_S, \; \displaystyle\int_0^{t_S} \frac{\left| \ddot{\theta_S}(t) \right|^2}{\overline{\ddot{\theta_S}}^2} \; dt \right]^\textrm{T} \\
        \textrm{s.t.} \; 
        & \max_{t\in[0,t_S]}\left| \dot{\theta_S}(t) \right| \leq \overline{\dot{\theta_S}} \\
        & \max_{t\in[0,t_S]}\left| \ddot{\theta_S}(t) \right| \leq \overline{\ddot{\theta_S}} \\
        & \max_{t\in[0,t_S]}\left| \alpha(t) \right| \leq \overline{\alpha} \\
        & \max_{t\in[0,t_S]}\left| \beta(t) \right| \leq \overline{\beta} \\
        & \theta_S(0) = \theta_{Si}, \; \dot{\theta_S}(0) = \ddot{\theta_S}(0) = \theta_S^{(3)}(0) = 0 \\
        & \theta_S(t_S) = \theta_{Sf}, \; \dot{\theta_S}(t_S) = \ddot{\theta_S}(t_S) = \theta_S^{(3)}(t_S) = 0 \\
        & \alpha(0) = \beta(0) = 0, \; \dot{\alpha}(0) = \dot{\beta}(0) = 0 \\
        & \alpha(t_S) = \beta(t_S) = 0, \; \dot{\alpha}(t_S) = \dot{\beta}(t_S) = 0
    \end{aligned}
    \label{eq:slewoptprob}
\end{equation}  

Due to the complicated non-linearly coupled payload dynamics during slewing, as represented by Eqs. (\ref{eq:slewalphasimp} - \ref{eq:slewbetasimp}), parameterization of the slew trajectory $\theta_S(t)$ in terms of $t_S$ is inadequate in transforming the above problem to a parameterized MOTOP. So, the differential flatness of the jib-payload system needs to be proved in order to conduct the slew trajectory planning via the flat output, which can be easily parameterized.

\subsubsection{Differential flatness of jib-payload system}
\label{subsubsec:flatoutputslew}

\begin{figure}[tbp]
     \begin{center}   
        \includegraphics[width=\columnwidth]{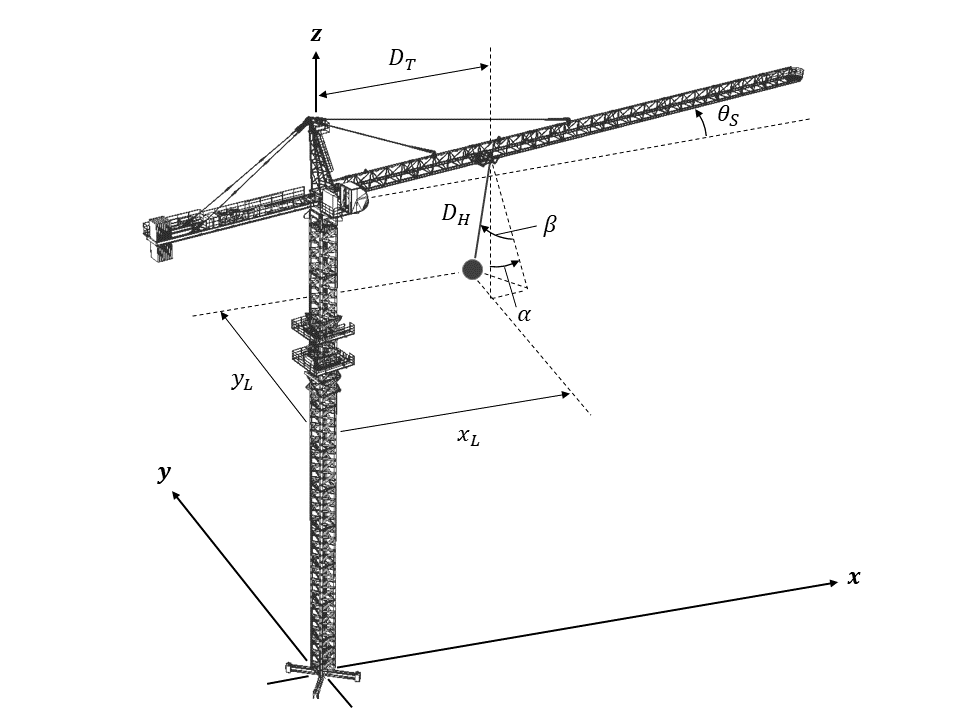}
        \caption{Auxiliary DOFs for the jib-payload system during slew operation.}
        \label{fig:slewauxDOFs}
    \end{center}
\end{figure}

Two auxiliary DOFs are considered for the slewing motion, which are the $x$ and $y$ co-ordinates of the payload during slew operation. The auxiliary DOFs are denoted by $x_L$ and $y_L$, as demonstrated in Figure \ref{fig:slewauxDOFs}.

From the geometric analysis of the slew motion, the auxiliary DOFs can be expressed in terms of the state variables as:

\begin{equation}
    x_L = D_T\cos{\theta_S} + D_H\cos{\beta}\sin{\alpha}\cos{\theta_S} - D_H\sin{\beta}\sin{\theta_S}
    \label{eq:slewauxDOFx}
\end{equation}

\begin{equation}
    y_L = D_T\sin{\theta_S} + D_H\cos{\beta}\sin{\alpha}\sin{\theta_S} + D_H\sin{\beta}\cos{\theta_S}
    \label{eq:slewauxDOFy}
\end{equation}  

Using the small angle approximation for the payload swing angles ($\sin{\alpha} \approx \alpha$, $\sin{\beta} \approx \beta$ and $\cos{\beta} \approx 1$):

\begin{equation}
    x_L = D_T\cos{\theta_S} + D_H\alpha\cos{\theta_S} - D_H\beta\sin{\theta_S}
    \label{eq:slewauxDOFxsimp}
\end{equation}

\begin{equation}
    y_L = D_T\sin{\theta_S} + D_H\alpha\sin{\theta_S} + D_H\beta\cos{\theta_S}
    \label{eq:slewauxDOFysimp}
\end{equation}  

Applying a few basic mathematical operations on the expressions of $x_L$ and $y_L$, it can be easily computed that:

\begin{equation}
    \alpha = \frac{1}{D_H} \left(x_L\cos{\theta_S} + y_L\sin{\theta_S} - D_T\right)
    \label{eq:slewalphathetas}
\end{equation}

\begin{equation}
    \beta = \frac{1}{D_H} \left(y_L\cos{\theta_S} - x_L\sin{\theta_S}\right)
    \label{eq:slewbetathetas}
\end{equation}  

As evident from the expressions of the swing angles above, if $\theta_S$ can be represented in terms of the auxiliary DOFs and a finite number of its time derivatives, all state variables are then expressed by the same. Then, $x_L$ and $y_L$ can be considered as the flat outputs of the system.

\begin{figure}[b]
     \begin{center}   
        \includegraphics[width=\columnwidth]{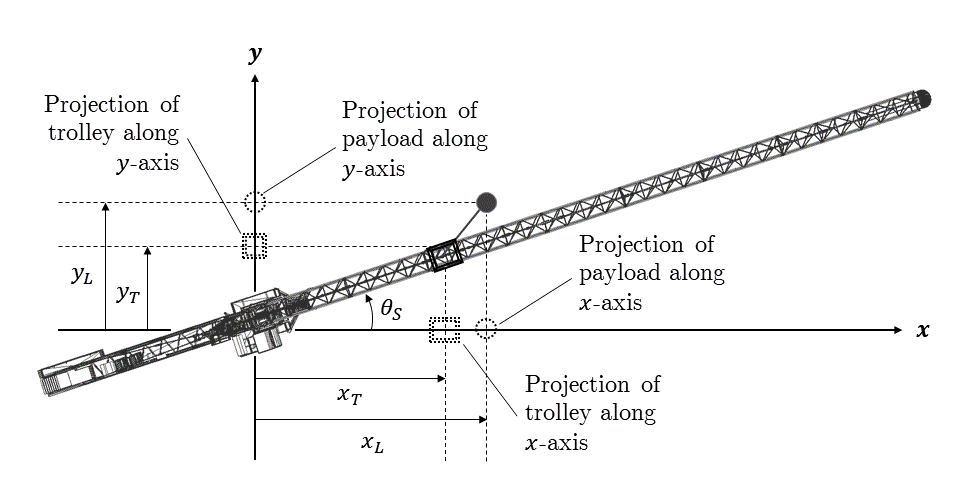}
        \caption{Position of the trolley and the payload during slew operation.}
        \label{fig:slewdynamicstop}
    \end{center}
\end{figure}

To find the relationship between the slew angle and the payload co-ordinates, the top view of the tower crane during slew operation is examined, as displayed in Figure \ref{fig:slewdynamicstop}. While the jib rotates on the $xy$ plane about the $z$ axis of the global co-ordinate system, the trolley, fixed on a location on the length of the jib, undergoes resulting linear translations along the $x$ and $y$ axis. This can be understood by considering the projection of the jib-trolley on the respective axes. Essentially, two simultaneous trolley displacements, $x_T$ and $y_T$, occur along two perpendicular directions during the pure slewing motion of the tower crane. Exploiting this fact, the slew operation is decomposed into two orthogonal trolley operations along the $x$ and the $y$ axis of the fixed reference system.

Now, for a trolley motion along a fixed direction, the trolley position and the payload position along that direction are related according to Eq. (\ref{eq:trolleydTfp}), derived in Section \ref{subsubsec:flatoutputtrolley}. Utilizing this association, the following two equations can be deduced for the two orthogonal trolley motions explained above:

\begin{equation}
    x_T = x_L + \frac{D_H}{g}\ddot{x_L} 
    \label{eq:slewxTfpx}
\end{equation}

\begin{equation}
    y_T = y_L + \frac{D_H}{g}\ddot{y_L} 
    \label{eq:slewyTfpy}
\end{equation}  

Knowing these positions as $x_T = D_T\cos{\theta_S}$ and $y_T = D_T\sin{\theta_S}$, the following relationships are calculated:

\begin{equation}
    \cos{\theta_S} = \frac{1}{D_T} \left(x_L + \frac{D_H}{g}\ddot{x_L}\right)
    \label{eq:slewthetaSfpx}
\end{equation}

\begin{equation}
    \sin{\theta_S} = \frac{1}{D_T} \left(y_L + \frac{D_H}{g}\ddot{y_L}\right)
    \label{eq:slewthetaSfpy}
\end{equation}  

So, the slew angle can be represented in terms of the two auxiliary DOFs during the slew motion as:

\begin{equation}
    \begin{aligned}
        \theta_S & = \arccos{\left[\frac{1}{D_T}\left(x_L + \frac{D_H}{g}\ddot{x_L}\right)\right]} \\
        & = \arcsin{\left[\frac{1}{D_T}\left(y_L + \frac{D_H}{g}\ddot{y_L}\right)\right]}
    \end{aligned}
    \label{eq:slewthetaSfp}
\end{equation}  

Then, substituting Eqs. (\ref{eq:slewthetaSfpx} - \ref{eq:slewthetaSfpy}) in Eqs. (\ref{eq:slewalphathetas} - \ref{eq:slewbetathetas}), the following expressions of the swing angles are computed:

\begin{equation}
    \alpha = \frac{1}{D_HD_T} \left[x_L^2 + y_L^2 + \frac{D_H}{g}\left(x_L\ddot{x_L} + y_L\ddot{y_L}\right) - D_T^2\right]
    \label{eq:slewalphafp}
\end{equation}

\begin{equation}
    \beta = \frac{1}{gD_T} \left(y_L\ddot{x_L} - x_L\ddot{y_L}\right)
    \label{eq:slewbetafp}
\end{equation}  

As indicated in the three equations above, the actuated DOF $\theta_S$, and the unactuated DOFs $\alpha$ and $\beta$ (and their velocities and accelerations via differentiation w.r.t. time), can be represented by the auxiliary DOFs $x_L$ and $y_L$, and some of their time derivatives. Therefore, it can be inferred that $x_L$ and $y_L$ are the flat outputs of the jib-payload system.

Differentiating Eq. (\ref{eq:slewthetaSfpx}) once w.r.t. time, and substituting $\sin{\theta_S}$ from Eq. (\ref{eq:slewthetaSfpy}), the following can be derived:

\begin{equation}
    \dot{\theta_S} = -\frac{\dot{x_L} + \frac{D_H}{g}x_L^{(3)}}{y_L + \frac{D_H}{g}\ddot{y_L}}
    \label{eq:slewthetaSvelfp}
\end{equation}  

Again, differentiating Eq. (\ref{eq:slewthetaSfpx}) twice w.r.t. time, and substituting $\sin{\theta_S}$ from Eq. (\ref{eq:slewthetaSfpy}), $\cos{\theta_S}$ from Eq. (\ref{eq:slewthetaSfpx}), and $\dot{\theta_S}$ from Eq. (\ref{eq:slewthetaSvelfp}), it is computed that:

\begin{equation}
    \ddot{\theta_S} = -\frac{x_L + \frac{D_H}{g}\ddot{x_L}}{y_L + \frac{D_H}{g}\ddot{y_L}} \left(\frac{\dot{x_L} + \frac{D_H}{g}x_L^{(3)}}{y_L + \frac{D_H}{g}\ddot{y_L}}\right)^2 - \frac{\ddot{x_L} + \frac{D_H}{g}x_L^{(4)}}{y_L + \frac{D_H}{g}\ddot{y_L}}
    \label{eq:slewthetaSaccfp}
\end{equation}  

Similarly, differentiating Eqs. (\ref{eq:slewalphafp} - \ref{eq:slewbetafp}) w.r.t. time once each, the velocities of the payload swing angles are obtained as:

\begin{equation}
    \begin{aligned}
        \dot{\alpha} = & \frac{1}{D_HD_T} \left(2x_L\dot{x_L} + 2y_L\dot{y_L}\right) \\
        & + \frac{1}{gD_T}\left(\dot{x_L}\ddot{x_L} + x_Lx_L^{(3)} + \dot{y_L}\ddot{y_L} + y_Ly_L^{(3)}\right)
    \end{aligned}
    \label{eq:slewalphavelfp}
\end{equation}

\begin{equation}
    \dot{\beta} = \frac{1}{gD_T} \left(\dot{y_L}\ddot{x_L} + y_Lx_L^{(3)} - \dot{x_L}\ddot{y_L} - x_Ly_L^{(3)}\right)
    \label{eq:slewbetavelfp}
\end{equation}  

Hence, all the state variables to be constrained during the slew operation are represented by the flat outputs.

\subsubsection{Slew trajectory planning in flat output space}
\label{subsubsec:slewtrajectoryflatoutput}

Since the tower crane has been shown as differentially flat during slew operation, the flat outputs can be put to use to plan the slew trajectory by planning the trajectories of the flat outputs. Thereby, the slew MOTOP presented through Eq. (\ref{eq:slewoptprob}) is prepared in terms of $x_L$ and $y_L$, and their time derivatives.

Based on the equality constraints presented in Eqs. (\ref{eq:sleweqconstinitial} - \ref{eq:slewswingeqconstfinal}), and the state variables given by Eqs. (\ref{eq:slewthetaSfp} - \ref{eq:slewbetavelfp}) in terms of the flat outputs, the following boundary constraints of $x_L$ and $y_L$ are deduced directly:

\begin{equation}
    \begin{aligned}
        & x_L(0) = D_T\cos{\theta_{Si}} = x_{Li}, \; \dot{x_L}(0) = 0, \; \ddot{x_L}(0) = 0, \\
        & x_L^{(3)}(0) = 0, \; x_L^{(4)}(0) = 0, \; x_L^{(5)}(0) = 0
    \end{aligned}
    \label{eq:slewfpxeqconstinitial}
\end{equation}

\begin{equation}
    \begin{aligned}
        & y_L(0) = D_T\sin{\theta_{Si}} = y_{Li}, \; \dot{y_L}(0) = 0, \; \ddot{y_L}(0) = 0, \\
        & y_L^{(3)}(0) = 0, \; y_L^{(4)}(0) = 0, \; y_L^{(5)}(0) = 0
    \end{aligned}
    \label{eq:slewfpyeqconstinitial}
\end{equation}

\begin{equation}
    \begin{aligned}
        & x_L(t_S) = D_T\cos{\theta_{Sf}} = x_{Lf}, \; \dot{x_L}(t_S) = 0, \; \ddot{x_L}(t_S) = 0, \\ 
        & x_L^{(3)}(t_S) = 0, \; x_L^{(4)}(t_S) = 0, \; x_L^{(5)}(t_S) = 0
    \end{aligned}
    \label{eq:slewfpxeqconstfinal}
\end{equation}

\begin{equation}
    \begin{aligned}
        & y_L(t_S) = D_T\sin{\theta_{Sf}} = y_{Lf}, \; \dot{y_L}(t_S) = 0, \; \ddot{y_L}(t_S) = 0, \\ 
        & y_L^{(3)}(t_S) = 0, \; y_L^{(4)}(t_S) = 0, \; y_L^{(5)}(t_S) = 0
    \end{aligned}
    \label{eq:slewfpyeqconstfinal}
\end{equation}  

Again, according to the inequality constraints for the slew motion in Eqs. (\ref{eq:slewineqconst} - \ref{eq:slewswingineqconst}), with the expressions of the state variables substituted:

\begin{equation}
    \begin{aligned}
        & \max_{t\in[0,t_S]} \, \left|\frac{\dot{x_L}(t) + \frac{D_H}{g}x_L^{(3)}(t)}{y_L(t) + \frac{D_H}{g}\ddot{y_L}(t)}\right| \leq \overline{\dot{\theta_S}}, \\
        & \max_{t\in[0,t_S]} \, \left| \frac{x_L(t) + \frac{D_H}{g}\ddot{x_L}(t)}{y_L(t) + \frac{D_H}{g}\ddot{y_L}(t)} \left(\frac{\dot{x_L}(t) + \frac{D_H}{g}x_L^{(3)}(t)}{y_L(t) + \frac{D_H}{g}\ddot{y_L}(t)}\right)^2\right| \\
        & + \max_{t\in[0,t_S]} \, \left| \frac{\ddot{x_L}(t) + \frac{D_H}{g}x_L^{(4)}(t)}{y_L(t) + \frac{D_H}{g}\ddot{y_L}(t)} \right| \leq \overline{\ddot{\theta_S}}
    \end{aligned}
    \label{eq:slewfpineqconstprelim}
\end{equation}

\begin{equation}
    \begin{aligned}
        & \max_{t\in[0,t_S]} \, \left|\frac{1}{D_HD_T} \left(x_L^2(t) + y_L^2(t)\right)\right| \\
        & + \max_{t\in[0,t_S]} \, \left|\frac{1}{gD_T}\left(x_L(t)\ddot{x_L}(t) + y_L(t)\ddot{y_L}(t)\right) - \frac{D_T}{D_H}\right| \leq \overline{\alpha}, \\
        & \max_{t\in[0,t_S]} \, \left|\frac{1}{gD_T} \left(y_L(t)\ddot{x_L}(t) - x_L(t)\ddot{y_L}(t)\right)\right| \leq \overline{\beta}
    \end{aligned}
    \label{eq:slewfpswingineqconstprelim}
\end{equation}  

Using the subadditivity property of absolute value inequality, the aforementioned equations can be written as:

\begin{equation}
    \begin{aligned}
        & \max_{t\in[0,t_S]} \, \left| \frac{\dot{x_L}(t) + \frac{D_H}{g}x_L^{(3)}(t)}{y_L(t) + \frac{D_H}{g}\ddot{y_L}(t)} \right| \leq \overline{\dot{\theta_S}}, \\
        & \max_{t\in[0,t_S]} \, \left| \frac{x_L(t) + \frac{D_H}{g}\ddot{x_L}(t)}{y_L(t) + \frac{D_H}{g}\ddot{y_L}(t)} \left(\overline{\dot{\theta_S}}^2 + \frac{\ddot{x_L}(t) + \frac{D_H}{g}x_L^{(4)}(t)}{x_L(t) + \frac{D_H}{g}\ddot{x_L}(t)}\right) \right| \\
        & \leq \overline{\ddot{\theta_S}}
    \end{aligned}
    \label{eq:slewfpineqconst}
\end{equation}

\begin{equation}
    \begin{aligned}
        & \max_{t\in[0,t_S]} \, \left| x_L^2(t) + y_L^2(t) + \frac{D_H}{g}\left(x_L(t)\ddot{x_L}(t) + y_L(t)\ddot{y_L}(t)\right) \right| \\
        & \leq D_T\left(D_H\overline{\alpha} - D_T\right), \\
        & \max_{t\in[0,t_S]} \, \left| \left(y_L(t)\ddot{x_L}(t) - x_L(t)\ddot{y_L}(t)\right) \right| \leq gD_T\overline{\beta}
    \end{aligned}
    \label{eq:slewfpswingineqconst}
\end{equation}  

Checking the 24 boundary conditions of the flat outputs (12 boundary conditions for each) given by Eqs. (\ref{eq:slewfpxeqconstinitial} - \ref{eq:slewfpyeqconstfinal}), 11\textsuperscript{th} degree B\'{e}zier curves are employed to parameterize $x_L(t)$ and $y_L(t)$. Following the parameterization, the flat outputs for the jib-payload system are determined in terms of $t_S$ as:

\begin{equation}
    x_L(t) = \displaystyle\sum_{r=0}^{11} \tensor[^r]{c}{_x}\left(\frac{t}{t_S}\right)^r; \qquad 0 \leq t \leq t_S
    \label{eq:slewfpxtrajectorypoly}
\end{equation}

\begin{equation}
    y_L(t) = \displaystyle\sum_{r=0}^{11} \tensor[^r]{c}{_y}\left(\frac{t}{t_S}\right)^r; \qquad 0 \leq t \leq t_S
    \label{eq:slewfpytrajectorypoly}
\end{equation}  

Writing the total change of the payload positions along the $x$ and $y$ directions (of the global reference system), during the slew motion, as $\Delta x_L = x_{Lf}-x_{Li}$ and $\Delta y_L = y_{Lf}-y_{Li}$, the coefficients $\tensor[^r]{c}{_x}$ and $\tensor[^r]{c}{_y}$ are derived as:

\begin{equation}
    \begin{aligned}
        & \tensor[^0]{c}{_x} = x_{Li}, \; \tensor[^1]{c}{_x} = 0, \; \tensor[^2]{c}{_x} = 0, \; \tensor[^3]{c}{_x} = 0, \; \tensor[^4]{c}{_x} = 0, \; \tensor[^5]{c}{_x} = 0, \\
        & \tensor[^6]{c}{_x} = 462\Delta x_L, \; \tensor[^7]{c}{_x} = -1980\Delta x_L, \; \tensor[^8]{c}{_x} = 3465\Delta x_L, \\
        & \tensor[^9]{c}{_x} = -3080\Delta x_L, \; \tensor[^{10}]{c}{_x} = 1386\Delta x_L, \; \tensor[^{11}]{c}{_x} = -252\Delta x_L
    \end{aligned}
    \label{eq:slewfpxtrajectorycoeff}
\end{equation}

\begin{equation}
    \begin{aligned}
        & \tensor[^0]{c}{_y} = y_{Li}, \; \tensor[^1]{c}{_y} = 0, \; \tensor[^2]{c}{_y} = 0, \; \tensor[^3]{c}{_y} = 0, \; \tensor[^4]{c}{_y} = 0, \; \tensor[^5]{c}{_y} = 0, \\
        & \tensor[^6]{c}{_y} = 462\Delta y_L, \; \tensor[^7]{c}{_y} = -1980\Delta y_L, \; \tensor[^8]{c}{_y} = 3465\Delta y_L, \\
        & \tensor[^9]{c}{_y} = -3080\Delta y_L, \; \tensor[^{10}]{c}{_y} = 1386\Delta y_L, \; \tensor[^{11}]{c}{_y} = -252\Delta y_L
    \end{aligned}
    \label{eq:slewfpytrajectorycoeff}
\end{equation}  

Similarly, the trajectories of the time derivatives of the flat outputs are also derived in their parameterized versions in terms of $t_S$, through consecutive differentiation of $x_L(t)$ and $y_L(t)$. Therefore, the slew trajectory optimization problem defined in Eq. (\ref{eq:slewoptprob}) is transformed to a MOTOP of the flat outputs, with only inequality constraints:

\begin{equation}
    \begin{aligned}
        \min_{t_S>0} \; & \left[ f_1(t_S), \; f_2(t_S) \right]^\textrm{T} \\
        \textrm{s.t.} \; & \textrm{Eqs. (\ref{eq:slewfpineqconst} - \ref{eq:slewfpswingineqconst})} \\
    \end{aligned}
    \label{eq:slewMOTOP}
\end{equation}

For Eq. (\ref{eq:slewMOTOP}), $x_L(t)$ and $y_L(t)$ are described according to Eqs. (\ref{eq:slewfpxtrajectorypoly} - \ref{eq:slewfpytrajectorypoly}), whose coefficients are calculated as Eqs. (\ref{eq:slewfpxtrajectorycoeff} - \ref{eq:slewfpytrajectorycoeff}). The solution to the slew MOTOP is the corresponding value of $t_S$ for the minimal slew time and energy. After $t_S$ is calculated, the optimal slew trajectories can be computed using Eqs. (\ref{eq:slewthetaSfp} - \ref{eq:slewthetaSaccfp}).

\section{Trajectory Optimization using Multi-objective Evolutionary Algorithm}
\label{sec:trajectoryoptimizationmoea}

For all the crane operations, the optimality criteria based on minimum time and energy is proposed in the current work to achieve optimal productivity by spending optimal effort. From a practical consideration, lower operating time is achieved by utilizing higher energy and lower operating energy is ensured at the expense of longer duration of operation. Since these two optimality criteria are conflicting in nature, the multi-objective optimization approach can only provide Pareto optimal solutions \cite{Miettinen1998}. For a multi-objective EA (MOEA), the two primary goals are to converge a set of non-dominated solutions towards the true Pareto front (known as PF\textsubscript{true}) and to maintain a diverse set of such solutions to represent a wide range of objective values for a \textit{posteriori} selection of an optimal solution. The Elitist Non-dominated Sorting Genetic Algorithm (NSGA-II) \cite{Deb2002} and the Generalized Differential Evolution 3 (GDE3) \cite{Kukkonena2005} are two well-established MOEA in the literature, well-known for their effective ability to achieve both the aforementioned goal with reduced computational complexity than other MOEAs. NSGA-II is a variant of Genetic Algorithm (GA) and GDE3 is a developed version of Differential Evolution (DE). Both these MOEAs employ a fast non-dominated sorting technique with elitism and an explicit diversity-preserving mechanism based on crowding distance sorting. The procedures with the generational loops of NSGA-II and GDE3 are listed in Algorithms \ref{algo:NSGAII} and \ref{algo:GDE3} respectively.

\begin{algorithm}
    \begin{algorithmic}[1]
        \STATE \textbf{procedure} $\mathcal{N}$ members evolve through $g$ generations to solve for $M$ objective functions
        \begin{ALC@g}
            \STATE Initialize population \textbf{P} of size $\mathcal{N}$ with randomly generated solutions;
            \STATE Evaluate solutions and assign rank (level) based on Pareto dominance - \textit{non-dominated sort};
            \STATE Generate offspring solution \textbf{Q} of size $\mathcal{N}$ using recombination and mutation operatosr;
            \FOR{$i$ = $1$ to $g-1$}
                \STATE Combine \textbf{P\textsubscript{$i$}} and \textbf{Q\textsubscript{$i$}} to generate population \textbf{R\textsubscript{$i$}} of size $2\mathcal{N}$;
                \STATE Evaluate solutions and assign rank (level) based on Pareto dominance to create fronts \textbf{F\textsubscript{$1$}} to \textbf{F\textsubscript{$k$}} - \textit{non-dominated sort};
                \STATE Initialize empty \textbf{P\textsubscript{$i+1$}} of size $\mathcal{N}$;
                \STATE Set $j$ = $1$;
                \WHILE{there is enough space in \textbf{P\textsubscript{$i+1$}} to add all solutions of \textbf{F\textsubscript{$j$}}}
                    \STATE Add members of \textbf{F\textsubscript{$j$}} to \textbf{P\textsubscript{$i+1$}};
                    \STATE Set $j$ = $j+1$;
                \ENDWHILE
                \STATE Calculate crowding distance of solutions in \textbf{F\textsubscript{$j$}} and sort in descending order - \textit{crowding distance sort};
                \STATE Fill the remaining spaces in \textbf{P\textsubscript{$i+1$}} by the corresponding number of best solutions from \textbf{F\textsubscript{$j$}};
                \STATE Generate offspring solution \textbf{Q\textsubscript{$i+1$}} of size $\mathcal{N}$ using recombination and mutation operators;
            \ENDFOR
        \end{ALC@g}
        \STATE \textbf{end procedure}    
    \end{algorithmic}
    \caption{NSGA-II}
    \label{algo:NSGAII}
\end{algorithm}

\begin{algorithm}
    \begin{algorithmic}[1]
        \STATE \textbf{procedure} $\mathcal{N}$ $D$-dimensional vectors evolve through $g$ generations to solve for $M$ objective functions
        \begin{ALC@g}
            \STATE Initialize population of size $\mathcal{N}$ with randomly generated solutions;
            \FOR{$i$ = $1$ to $g-1$}
                \FOR{$k$ = $1$ to $\mathcal{N}$}
                    \STATE Randomly select three mutually different parent solutions $\textbf{x}_{r1}$, $\textbf{x}_{r2}$ and $\textbf{x}_{r3}$ and a random variable index $j_{rand} \in \{1,...,D\}$;
                    \STATE Find $\textbf{u} = \textbf{x}_{r3} + F(\textbf{x}_{r1} - \textbf{x}_{r2})$ with scaling factor $F$;
                    \FOR{$j$ = $1$ to $D$}
                        \IF{$rand_j[0,1] < CR$ OR $j = j_{rand}$ with crossover rate CR}
                            \STATE $u_{j,k,i} = u_{j,k,i}$;
                        \ELSE
                            \STATE $u_{j,k,i} = x_{j,k,i}$;
                        \ENDIF
                    \ENDFOR
                    \STATE Evaluate fitness of all the child solutions $\textbf{u}$;
                    \IF{$\textbf{u}$ and $\textbf{x}_k$ of the parent population are indifferent to each other}
                        \STATE Add both solutions to the offspring population;
                    \ELSIF{$\textbf{x}_k$ dominates $\textbf{u}$}
                        \STATE Add $\textbf{x}_k$ to the offspring population;
                    \ELSE
                        \STATE Add $\textbf{u}$ to the offspring population;
                    \ENDIF
                \ENDFOR
                \STATE Evaluate solutions and assign rank - \textit{non-dominated sort};
                \STATE Generate next generation with the $\mathcal{N}$ best solutions;
            \ENDFOR
        \end{ALC@g}
        \STATE \textbf{end procedure}    
    \end{algorithmic}
    \caption{GDE3}
    \label{algo:GDE3}
\end{algorithm}

In the present work, to solve the MOTOPs presented via Eqs. (\ref{eq:hoistMOTOP}), (\ref{eq:trolleyMOTOP}) and (\ref{eq:slewMOTOP}), both NSGA-II and GDE3 are applied to generate Pareto optimal solutions. The MOEA more suitable for the MOTOPs discussed in Section \ref{sec:trajectoryoptimizationproblem} is to be selected based on certain performance metrics.

\subsection{Performance Metrics for MOEAs}
\label{subsec:metricsforMOEAs}

To analyze the performances of the two MOEAs, when applied to the MOTOP, the required scales of comparison need to consider the general objectives of multi-objective optimization that should be realized by the MOEA in the course of finding a Pareto optimal set.

\begin{enumerate}
    \item The PF\textsubscript{known} should be as close to PF\textsubscript{true}. This is to ensure that the Pareto solutions found by the MOEA are convergent towards to the best possible non-dominated solutions.
    \item The members of the Pareto set at the end of the MOEA cycle should cover as much of the feasible objective space as possible. So, the spread of the solutions should be high enough for the user to get a wide range of candidate solutions, out of which one can be used according to preference or problem-domain knowledge.
    \item As MOTOP represents a real-world problem where the crane operation trajectories are generated to be used as reference inputs for the tracking controller, the optimal results should be obtained by the MOEA as fast as possible. This also helps in employing fast re-planned trajectories in near real-time, in dynamic environments.
\end{enumerate}

To address these factors, the following metrics are considered in the current study to select an MOEA for the tower crane trajectory planner: hyperarea \cite{Coello2007}, spacing \cite{Coello2007} and runtime. While a higher hyperarea for the solutions from an MOEA would denote a Pareto front closer to the PF\textsubscript{true}, a lower spacing for a Pareto set indicates a more even spread of solutions over the objective space. Therefore, the MOEA with a higher hyperarea, lower spacing, and less runtime at the end of a specific number of function evaluations is sought for the MOTOP. The simulation studies done in Section \ref{subsec:moeafortrajectoryoptimization} employ these metrics to analyze the two MOEAs presented in this section.

\subsection{Selection of Optimal Solution from Pareto Set}
\label{subsec:AFMFselection}

To decide on a single solution from the set of Pareto optimal solutions provided by the chosen MOEA, the objectives are weighted according to the preference of the decision-maker, or using some problem-domain knowledge of the user. However, this weighting depends on the particular application and other environmental conditions. As a result, the preference is seldom fixed and varies in a fuzzy manner based on the scenario of the problem. For the MOTOP discussed in this research, a similar fuzzy behavior of the relative ranking of the objectives (operating time and effort) is expected. To deal with this, a mean linear fuzzy membership function is employed to characterize the fuzzy goals of the user.

For a set of Pareto optimal solutions obtained by solving the MOTOP, the solutions range from a minimum of individual objective values to their maximum. A linear fuzzy membership function for each objective function value of a solution $x_i$ is constructed as:

\begin{equation}
    \mu_1(x_i) = \frac{f_{1max}-f_1(x_i)}{f_{1max}-f_{1min}}
    \label{eq:mu1}
\end{equation}

\begin{equation}
    \mu_2(x_i) = \frac{f_{2max}-f_2(x_i)}{f_{2max}-f_{2min}}
    \label{eq:mu2}
\end{equation}

$\mu_1(x_i)$ and $\mu_2(x_i)$ attain the value 1 when $f_{1min}$ and $f_{2min}$ is reached, respectively. For the bi-objective minimization problem of the study, both these cases cannot be attained at the same time, due to the conflicting nature of the objectives. Therefore, the mean of the linear fuzzy membership functions is calculated as a metric to pick a solution from the set.

\begin{equation}
    \overline{\mu}(x_i) = \frac{\mu_1(x_i)+\mu_2(x_i)}{2}
    \label{eq:mubar}
\end{equation}

From this function, the solution with a higher value of $\overline{\mu}$ is considered to be more suitable as it represents higher weighting of both objectives, albeit in a fuzzy way. Hence, the solution from the Pareto set with $\overline{\mu}_{max}$ is selected as an optimal solution to the MOTOP.

\section{Anti-swing Trajectory Planning of a Lifting Path}
\label{sec:liftingtrajectory}

The optimal lifting path planned by the path planning module of the CALP system \cite{Cai2016,Huang2018} is represented as an ordered array of configurations (operation switching points or way-points), which stands for a general poly-line in the Cartesian space. Since, the operations of the tower crane are decoupled, portions of the poly-line are axis-aligned. As a result, the decoupled lifting path P of the tower crane is defined in the C-space as:

\begin{equation}
    P = \{(p_j, t_j)\}_{j=0,1,2,...,n} \cup \{a_k\}_{k=1,2,3,...,n}
\end{equation}
\label{eqn:liftingpath}

In this definition, $p_j=[\theta_S(t_j), d_T(t_j), d_H(t_j)]^T$ is the $j$-th way-point with time instant $t_j$. It is to be noted that $p_0$ and $p_n$ denote the start and the end configurations of the lifting path. Without any loss of generality, it can be established that $t_0=0<t_1<t_2<t_3<\dots<t_n$. The $k$-th actuated operation is denoted by $a_k$ occurring in $[(p_{k-1}, t_{k-1}), (p_k, t_k)]$. Total number of actuated operations $n$ for the lifting path is denoted by:

\begin{equation}
    n = n_{hoist}+n_{trolley}+n_{slew}
\end{equation}
\label{eqn:operations}

Here, $n_{hoist}$, $n_{trolley}$ and $n_{slew}$ are the number of hoist, trolley and slew operations, respectively, in the lifting path.

Each operation $a_k$ between two way-points $(p_{k-1}, t_{k-1})$ and $(p_k, t_k)$ is executed by any one of the actuators whose trajectory profile determine the position, velocity, acceleration and jerk of the concerned operation. Hence, the individual one-dimensional trajectories of the three actuators ($\theta_S(t)$, $d_T(t)$ and $d_H(t)$) are the ordered sum of trajectories between specific way-points. In this case, the optimal position, velocity and acceleration profiles of each operation are first locally interpolated between a pair of way-points, satisfying the motion constraints and eliminating the residual payload swing, according to the methods presented in Sections \ref{sec:trajectoryoptimizationproblem} and \ref{sec:trajectoryoptimizationmoea}. The optimal planning of all the operation trajectories can be performed in parallel, as they are decoupled. The anti-swing trajectory planning time, $t_{plan}$, is the maximum runtime of the MOEA-based optimizer in solving the MOTOP of the individual operations. Then the trajectories are assigned to the controller of the corresponding actuator executing the operation. Duration of all the operations planned are added cumulatively to find ${t_1,...,t_n}$. Finally, the ordered sequences of optimal operation trajectories for each actuator are combined to obtain the trajectory of that particular actuator during a lifting task.

\begin{figure}[tbp]
    \begin{center}
        \includegraphics[width=\columnwidth]{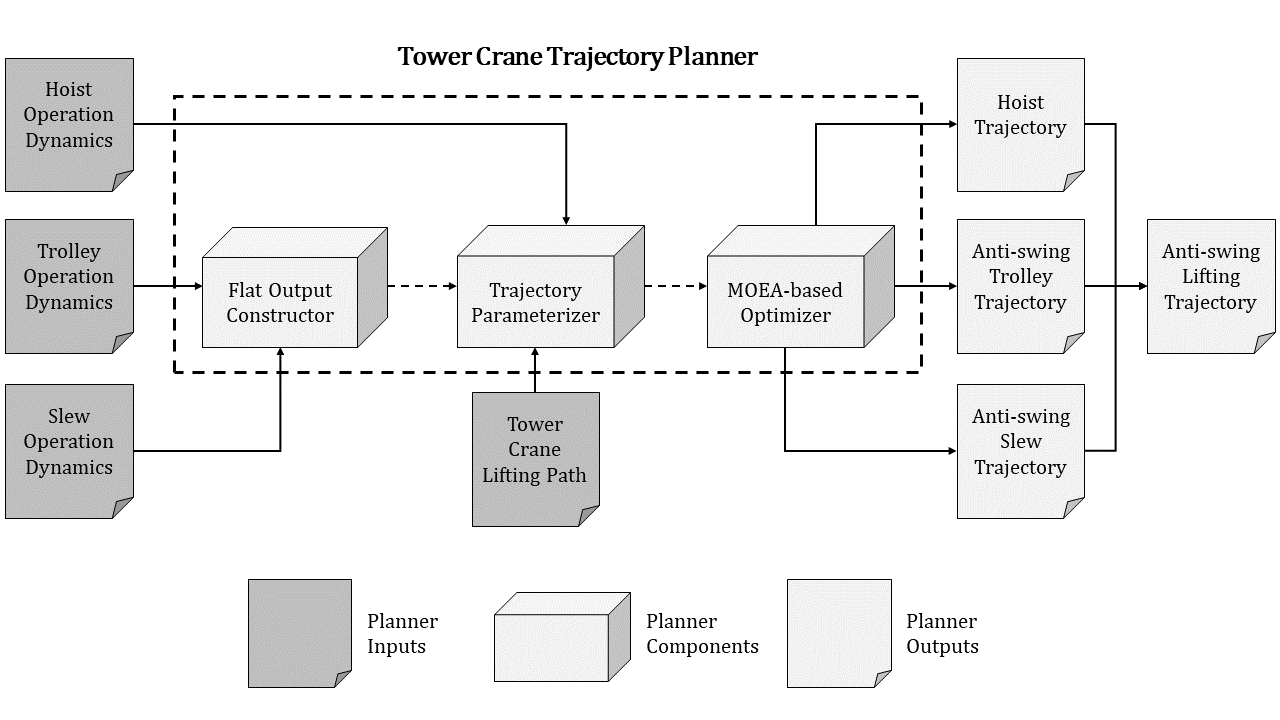}
        \caption{Anti-swing tower crane trajectory planning architecture for the CALP system.}
        \label{fig:trajectoryplanner}
    \end{center}
\end{figure}

The overall anti-swing tower crane lifting trajectory planning architecture for the CALP system is demonstrated in Figure \ref{fig:trajectoryplanner}.

\section{Simulation Studies}
\label{sec:casestudies}

\subsection{System Specifications}
\label{subsec:systemspecifications}

The simulation studies are conducted in MATLAB R2021a environment, on a PC with a 2.20 GHz Intel(R) Core(TM) i7-8750H CPU and 8 GB installed physical memory (RAM), run on Microsoft Windows 10 operating system. PlatEMO v3.2 \cite{Tian2017} is used to employ the NSGA-II and GDE3 algorithms. An NVIDIA GeForce GTX 1650 display adapter with 896 NVIDIA CUDA cores and 4 GB standard memory configuration is utilized as the GPU.

\subsection{Selection of MOEA for Trajectory Optimization}
\label{subsec:moeafortrajectoryoptimization}

\subsubsection{Procedure for MOEA comparison}
\label{subsec:MOEAaprocedure}

NSGA-II and GDE3 are applied to the multi-objective trajectory planning problem of all three crane operations to analyze and compare the performances of the MOEAs over multiple applications. Experiments are done for a 1:10 scaled virtual model of the Terex SK 415-20 hammerhead tower crane. Limits on the tower crane motion given in Table \ref{tbl:cranelimits} are utilized to perform trajectory planning for the operations listed in Table \ref{tbl:operationsMOEA}. The operational values (boundary positions) are obtained from the lifting path planner of the CALP system, as portions of a lifting task. The maximum swing angle values are selected based on the maximum allowed deflection of the payload from the planned lifting path, according to the threshold distance given by the multi-level OBBs of the payload. The values are also kept within the small-angle approximation ranges ($< 5^\circ$). The gravitational acceleration $g$ is taken as 9.8m/s\textsuperscript2.

\begin{table}[tbp]
    \centering
    \caption{Mechanical constraints and safety limits of the Terex SK 415-20 hammerhead tower crane used in the simulation studies.}
    \begin{tabular}{lc}
        \hline
        \bf Parameter & \bf Value \\
        \hline
        minimum hoist velocity ($\underline{\dot{d_H}}$) & 0.1m/s \\
        maximum hoist velocity ($\overline{\dot{d_H}}$) & 0.3m/s \\
        maximum hoist acceleration ($\overline{\ddot{d_H}}$) & 0.2m/s\textsuperscript2 \\
        minimum trolley velocity ($\underline{\dot{d_T}}$) & 0.05m/s \\
        maximum trolley velocity ($\overline{\dot{d_T}}$) & 0.25m/s \\
        maximum trolley acceleration ($\overline{\ddot{d_T}}$) & 0.2m/s\textsuperscript2 \\
        minimum slew velocity ($\underline{\dot{\theta_S}}$) & 3$^\circ$/s \\
        maximum slew velocity ($\overline{\dot{\theta_S}}$) & 10$^\circ$/s \\
        maximum slew acceleration ($\overline{\ddot{\theta_S}}$) & 10$^\circ$/s\textsuperscript2 \\
        maximum radial payload swing angle ($\overline{\alpha}$) & 2.5$^\circ$ \\
        maximum tangential payload swing angle ($\overline{\beta}$) & 2.5$^\circ$ \\
        \hline
    \end{tabular}
    \label{tbl:cranelimits}
\end{table}

\begin{table}[tbp]
    \centering
    \caption{Tower crane operations in the experiments for MOEA comparison.}
    \begin{tabular}{llc}
        \hline
        \bf Operation & \bf Parameter & \bf Value \\
        \hline
        \multirow{2}{*}{Hoist} & initial hoisting height ($d_{Hi}$) & 5m \\
        & final hoisting height ($d_{Hf}$) & 4m \\
        \hline
        \multirow{3}{*}{Trolley} & initial trolley position ($d_{Ti}$) & 2m \\
        & final trolley position ($d_{Tf}$) & 2.5m \\
        & fixed length of hoist-cable ($D_H$) & 5m \\
        \hline
        \multirow{4}{*}{Slew} & initial slew angle ($\theta_{Si}$) & 50$^\circ$ \\
        & final slew angle ($\theta_{Sf}$) & 80$^\circ$ \\
        & fixed length of hoist-cable ($D_H$) & 5m \\
        & fixed position of trolley ($D_T$) & 2.5m \\
        \hline
    \end{tabular}
    \label{tbl:operationsMOEA}
\end{table}

The implementations of both the MOEAs are configured with the parameters shown in Table \ref{tbl:parametersMOEA}. The operator values are obtained according to the proposals by the respective authors. Minor tuning was conducted for the mutation probability of NSGA-II and the crossover rate of GDE3. NSGA-II is run via a real number-based genetic encoding scheme. Both the MOEAs are initialized by generating a population with solutions created according to a uniform random distribution spanning the feasible interval of the decision variable space. The upper bound of the decision variable space was calculated assuming continuous-velocity operations executed with their respective minimum velocities.

\begin{table}[tbp]
    \centering
    \caption{Parameter configurations for NSGA-II and GDE3 in the experiments for MOEA comparison.}
    \begin{tabular}{llc}
        \hline
        \bf Algorithm & \bf Parameter & \bf Value \\
        \hline
        \multirow{4}{*}{NSGA-II} & crossover probability & 0.9 \\
        & crossover distribution index & 20 \\
        & mutation probability & 0.5 \\
        & mutation distribution index & 20 \\
        \hline
        \multirow{2}{*}{GDE3} & crossover rate & 0.9 \\
        & scaling factor & 0.5 \\
        \hline
        \multirow{3}{*}{Both} & population size & 100 \\
        & maximum function evaluations & 5000 \\
        & feasible decision variable space & [0,10] \\
        \hline
    \end{tabular}
    \label{tbl:parametersMOEA}
\end{table}

\subsubsection{Simulated MOEA results and analysis}
\label{subsec:MOEAresults}

\begin{figure}[tbp]
    \begin{center}
        \includegraphics[width=\columnwidth]{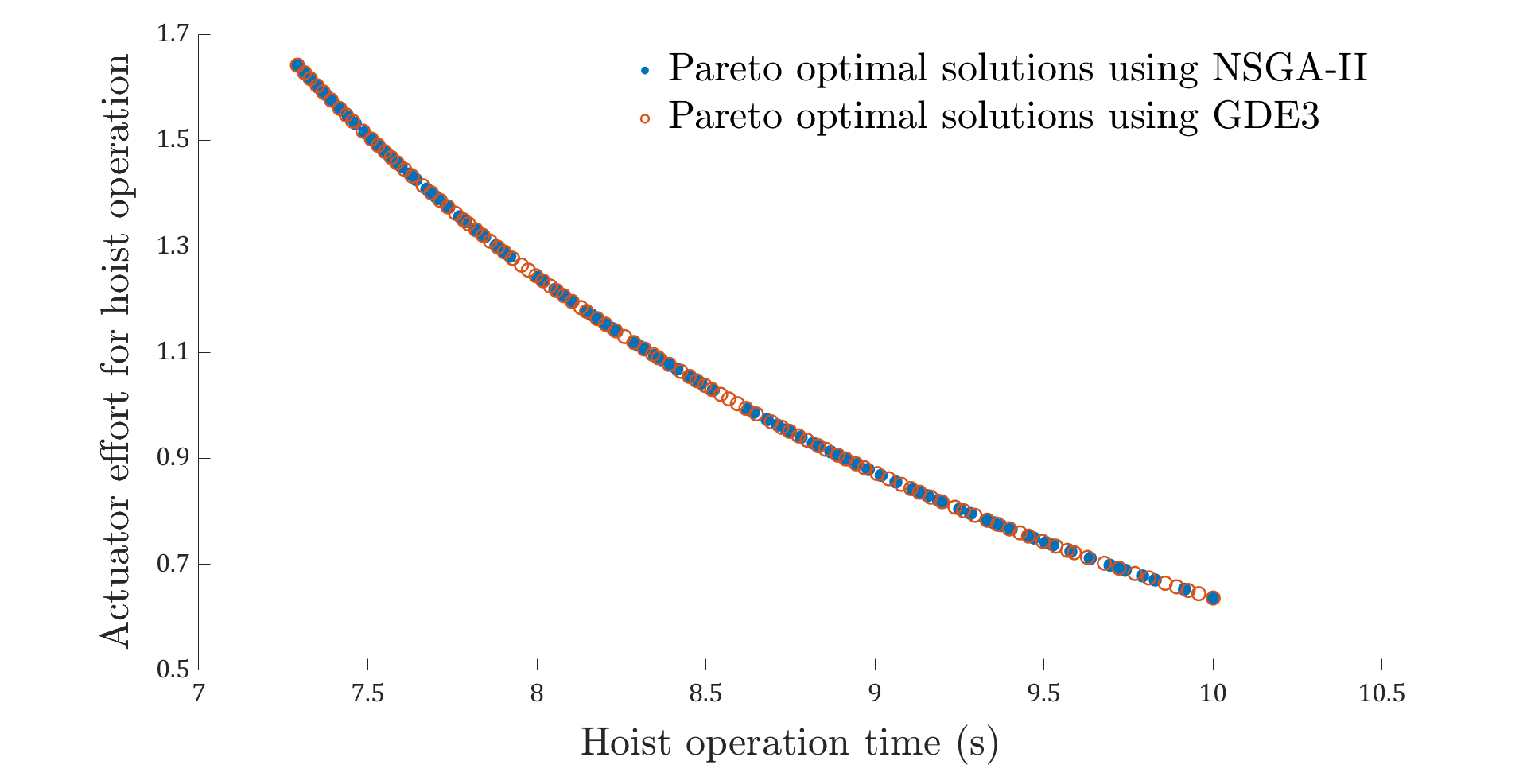}
        \caption{Pareto optimal solutions by NSGA-II and GDE3 for the MOTOP of the hoist operation in Table \ref{tbl:operationsMOEA} subjected to constraints in Table \ref{tbl:cranelimits}.}
        \label{fig:hoistMOEA}
    \end{center}
\end{figure}

\begin{figure}[tbp]
    \begin{center}
        \includegraphics[width=\columnwidth]{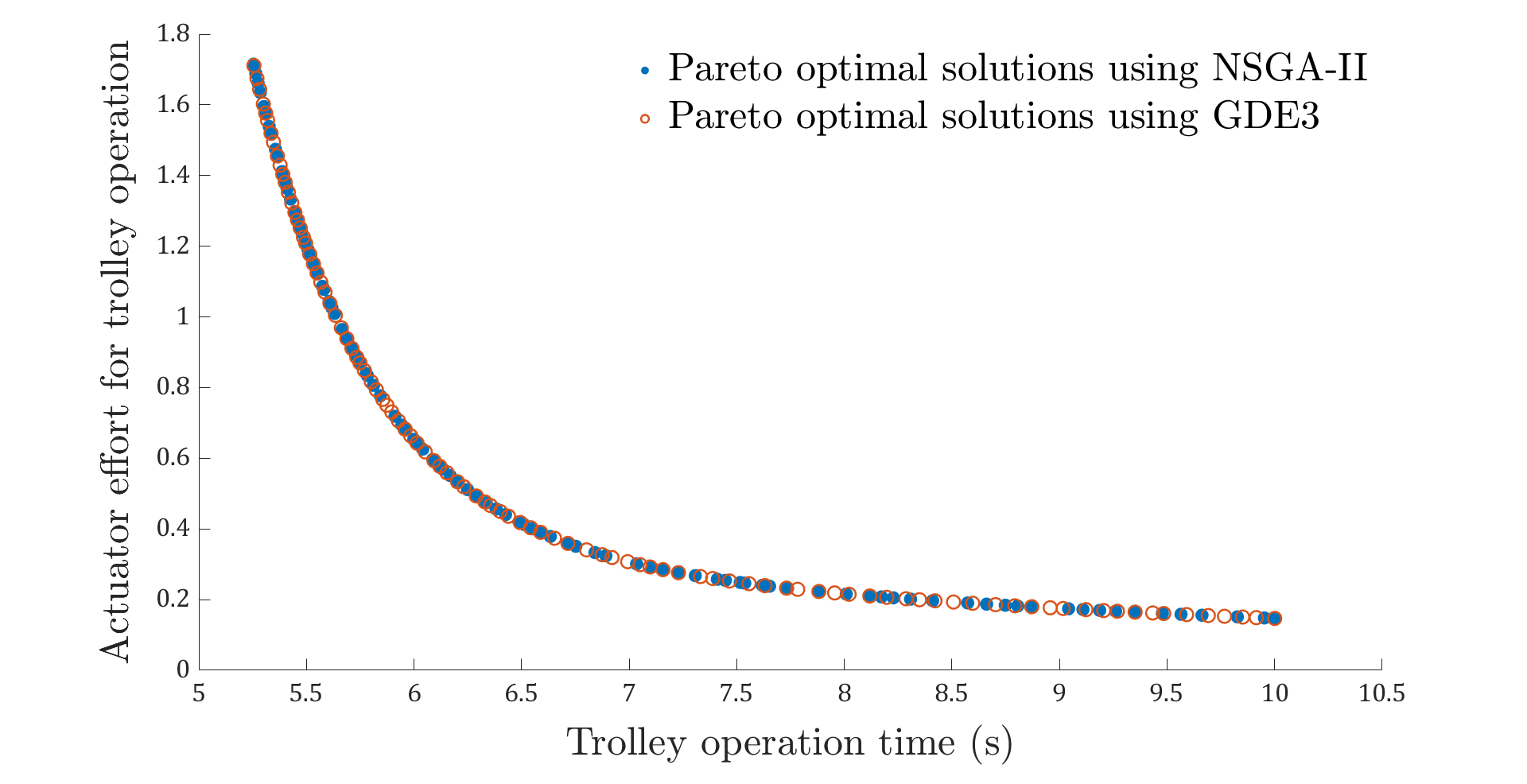}
        \caption{Pareto optimal solutions by NSGA-II and GDE3 for the MOTOP of the trolley operation in Table \ref{tbl:operationsMOEA} subjected to constraints in Table \ref{tbl:cranelimits}.}
        \label{fig:trolleyMOEA}
    \end{center}
\end{figure}

\begin{figure}[tbp]
    \begin{center}
        \includegraphics[width=\columnwidth]{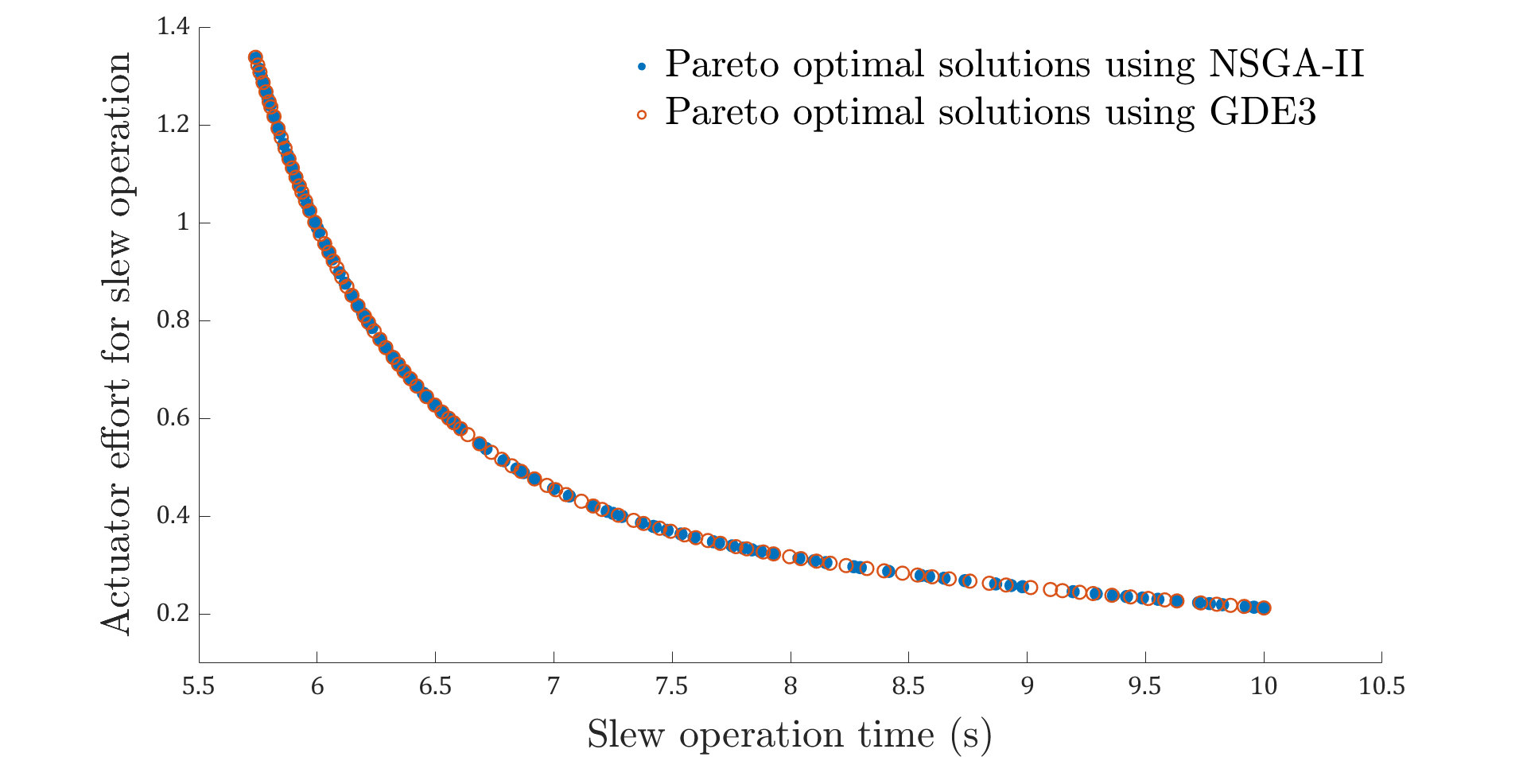}
        \caption{Pareto optimal solutions by NSGA-II and GDE3 for the MOTOP of the slew operation in Table \ref{tbl:operationsMOEA} subjected to constraints in Table \ref{tbl:cranelimits}.}
        \label{fig:slewMOEA}
    \end{center}
\end{figure}

For the hoist, trolley and slew operations, the best values of operating time and actuator effort satisfying all the constraints, obtained by the MOEAs are illustrated in Figures \ref{fig:hoistMOEA}, \ref{fig:trolleyMOEA}, \ref{fig:slewMOEA}, respectively. As can be seen, both NSGA-II and GDE3 have converged almost to the same front for all three operations. So, given an equal number of solutions in the initial population and the same maximum number of function evaluations, both MOEAs exhibit almost equal convergence.

Table \ref{tbl:optimalsMOEA} lists the bounds of the Pareto fronts found by the two MOEA, ($f_{1min}$,$f_{2max}$) and ($f_{1max}$,$f_{2min}$), in the feasible objective space, for the three crane operations. The maximum values of the mean linear fuzzy membership function $\overline{\mu}_{max}$ for those fronts, and the corresponding selected optimal solution, are also presented in the aforementioned tables. The identical values for these parameters further demonstrate the immense competition between NSGA-II and GDE3. Comparison based on just these values puts both algorithms as a candidate for providing the optimal solution to the MOTOP for any of the tower crane operations. To further investigate their performance, the metrics discussed in Section \ref{subsec:metricsforMOEAs} are examined. 

\begin{table}[tbp]
    \centering
    \caption{Optimal solutions provided by NSGA-II and GDE3 for the MOTOP of the tower crane operations in Table \ref{tbl:operationsMOEA} subjected to constraints in Table \ref{tbl:cranelimits}.}
    \begin{tabular}{lp{0.3\columnwidth}cc}
        \hline
        \bf Operation & \bf Parameter & \bf NSGA-II & \bf GDE3 \\
        \hline
        \multirow{4}{*}{Hoist} & min time, max effort ($f_{1min}$,$f_{2max}$) & (7.29,1.64) & (7.29,1.64) \\
        & max time, min effort ($f_{1max}$,$f_{2min}$) & (10,0.64) & (10,0.64) \\
        & max mean fuzzy membership $\overline{\mu}_{max}$ & 0.58 & 0.58 \\
        & Optimal solution ($f_{1opt},f_{2opt}$) & (8.48,1.04) & (8.48,1.04) \\
        \hline
        \multirow{4}{*}{Trolley} & min time, max effort ($f_{1min}$,$f_{2max}$) & (5.25,1.72) & (5.26,1.71) \\
        & max time, min effort ($f_{1max}$,$f_{2min}$) & (10,0.15) & (10,0.15) \\
        & max mean fuzzy membership $\overline{\mu}_{max}$ & 0.78 & 0.78 \\
        & Optimal solution ($f_{1opt},f_{2opt}$) & (6.49,0.42) & (6.44,0.44) \\
        \hline
        \multirow{4}{*}{Slew} & min time, max effort ($f_{1min}$,$f_{2max}$) & (5.74,1.34) & (5.74,1.34) \\
        & max time, min effort ($f_{1max}$,$f_{2min}$) & (10,0.21) & (10,0.21) \\
        & max mean fuzzy membership $\overline{\mu}_{max}$ & 0.74 & 0.74 \\
        & Optimal solution ($f_{1opt},f_{2opt}$) & (6.91,0.48) & (6.92,0.48) \\
        \hline
    \end{tabular}
    \label{tbl:optimalsMOEA}
\end{table}

The computed performance metrics of the two MOEAs for trajectory optimization of all three crane operations are compiled in Table \ref{tbl:metricsMOEA}. NSGA-II is proven as the faster method since it ranks always higher w.r.t. $T_{run}$. On average, it takes around 15.48\% less time to compute a set of Pareto optimal solutions using a given number of function evaluations, than GDE3. This makes NSGA-II the preferred solution approach for conditions requiring faster planning, such as trajectory re-planning in dynamic autonomous construction environments. The reference trajectory input to the crane controller can be provided in real-time or near real-time, based on the results of the runtime of the algorithm. However, GDE3 performs significantly better than NSGA-II in terms of $S$, which is also evident from the Pareto fronts in Figures \ref{fig:hoistMOEA}, \ref{fig:trolleyMOEA} and \ref{fig:slewMOEA}. The Pareto set found by GDE3 is always more evenly spaced throughout the feasible objective space, as indicated by its lower $S$ value. The average improvement in the spread of the solutions of GDE3 compared to that of NSGA-II, is around 33\%. Hence, the output from GDE3 can be considered as a superior technique for solving the MOTOP of tower crane operations, when the decision-maker wants to explore the objective space within a diverse range, complying to different construction scenarios. Moreover, the hyperarea covered by the PF\textsubscript{known} of GDE3 is consistently closer to the PF\textsubscript{true}, denoted by its higher, although marginally, $HA$ values than its NSGA-II counterpart. Since the two algorithms converge on almost the same front for all crane operations, this improvement might not be significant.

\begin{table}[tbp]
    \centering
    \caption{Performance metrics of NSGA-II and GDE3 for solving the MOTOP of the tower crane operations in Table \ref{tbl:operationsMOEA} subjected to constraints in Table \ref{tbl:cranelimits}. (Better metric values of an MOEA are highlighted in bold, accompanied by the percentage improvement compared to that of the other method.)}
    \begin{tabular}{llcc}
        \hline
        \bf Operation & \bf Metric & \bf NSGA-II & \bf GDE3 \\
        \hline
        \multirow{3}{*}{Hoist} & runtime ($T_{run}$) & \textbf{0.77s} (+15.38\%) & 0.91s \\
        & spacing ($S$) & 0.015 & \textbf{0.008} (+46.67\%) \\
        & hyperarea ($HA$) & 0.334 & \textbf{0.336} (+0.60\%) \\
        \hline
        \multirow{3}{*}{Trolley} & runtime ($T_{run}$) & \textbf{0.99s} (+20.80\%) & 1.25s \\
        & spacing ($S$) & 0.027 & \textbf{0.019} (+29.63\%) \\
        & hyperarea ($HA$) & 0.512 & \textbf{0.521} (+1.75\%) \\
        \hline
        \multirow{3}{*}{Slew} & runtime ($T_{run}$) & \textbf{2.01s} (+10.27\%) & 2.24s \\
        & spacing ($S$) & 0.022 & \textbf{0.017} (+22.72\%) \\
        & hyperarea ($HA$) & 0.459 & \textbf{0.466} (+1.52\%) \\
        \hline
    \end{tabular}
    \label{tbl:metricsMOEA}
\end{table}

Eventually, considering the statistical upper hand of the metrics of one MOEA over the other, GDE3 is selected as the optimizer for the trajectory optimization problem of tower crane operations. This is to facilitate the fast generation of diverse Pareto optimal solutions as close to the true Pareto front as possible. The obtained solutions can then be evaluated based on the mean linear fuzzy membership function, to compute a single optimal solution for the corresponding MOTOP.

\subsection{Anti-swing Lifting Trajectory Planning}
\label{subsec:liftingtrajectoryplanning}

Based on the aforementioned analysis and selection of GDE3 as the MOEA for the MOTOP of the tower crane operations, a complete anti-swing trajectory for a tower crane lifting path can be planned. This section employs the developed anti-swing trajectory planner of the CALP system, with the functional architecture demonstrated in Figure \ref{fig:trajectoryplanner}. GDE3 acts as the MOEA-based solver of the trajectory planner.

\subsubsection{Procedure for trajectory planning}
\label{subsec:ltpprocedure}

\begin{figure}[tbp]
    \begin{center}
        \includegraphics[width=\columnwidth]{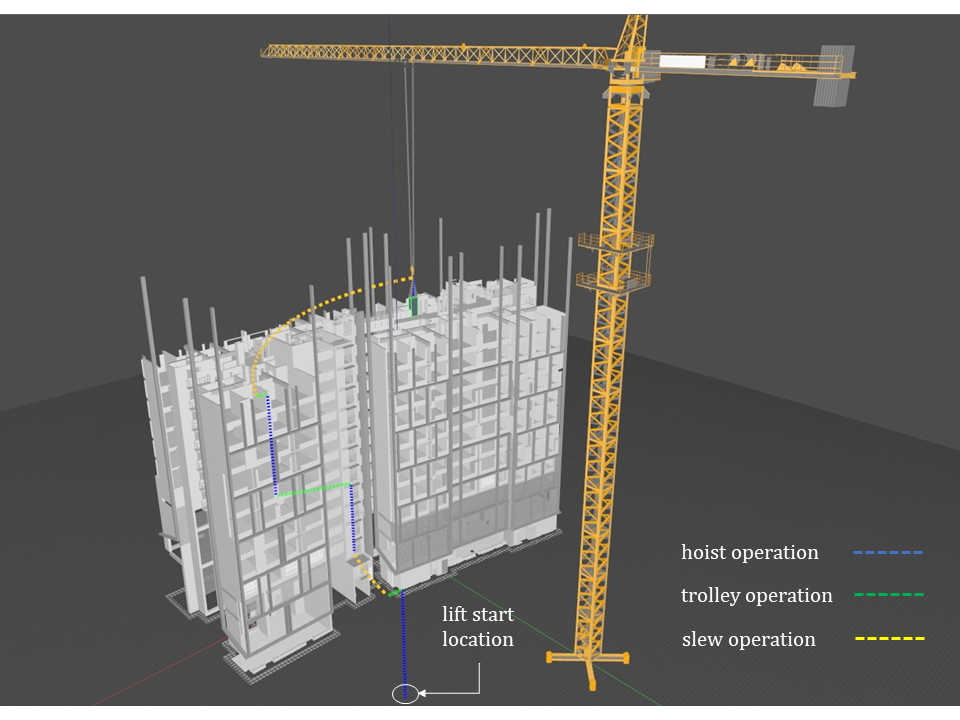}
        \caption{Tower crane lifting path from the path planner of the CALP system used in the trajectory planning simulation study.}
        \label{fig:plannedliftingpath}
    \end{center}
\end{figure}

An optimal collision-free lifting path for a building component is planned using the in-house path planner of the CALP system \cite{Cai2016,Dutta2020}. A progressive BIM model of an in-construction residential building and a 3D model of the Terex SK 415-20 hammerhead tower crane, both with a scale 1:10, is utilized to represent the autonomous construction scene. The lifting path is illustrated in Figure \ref{fig:plannedliftingpath}, consisting of the fundamental crane operations.

The crane configurations at the way-points of the lifting path are extracted and listed in Table \ref{tbl:decoupledliftingpath} with corresponding time instants. Based on this, the order of crane operations is compiled with corresponding parameters (initial and final positions, fixed DOFs) during each motion, as demonstrated in Table \ref{tbl:liftingoperations}. In this lifting path, $n_{hoist}$ = 4, $n_{trolley}$ = 3, and $n_{slew}$ = 2. For each of the 9 crane operations, the MOTOP is solved using GDE3. The parameter configurations provided in Table \ref{tbl:parametersMOEA} are used for the simulation study. Limits on the tower crane motion given in Table \ref{tbl:cranelimits} are utilized to perform the trajectory optimization. The gravitational acceleration $g$ is taken as 9.8m/s\textsuperscript2.
        
\begin{table}[tbp]
    \centering
    \caption{Tower crane lifting configurations with associated time instants of the lifting path used in the trajectory planning simulation study.}
    \begin{tabular}{>{\centering\arraybackslash}m{0.2\columnwidth}>{\centering\arraybackslash}m{0.2\columnwidth}>{\centering\arraybackslash}m{0.2\columnwidth}>{\centering\arraybackslash}m{0.2\columnwidth}}
        \hline
        \bf Slew angle ($^\circ$) & \bf Trolley position (m) & \bf Hoisting height (m) & \bf Time instant \\
        \hline
        83 & 2.4 & 5.5 & $t_0 = 0$ \\
        83 & 2.4 & 4.1 & $t_1$ \\
        83 & 2.6 & 4.1 & $t_2$ \\
        66 & 2.6 & 4.1 & $t_3$ \\
        66 & 2.6 & 3.3 & $t_4$ \\
        66 & 3.3 & 3.3 & $t_5$ \\
        66 & 3.3 & 2.2 & $t_6$ \\
        66 & 3.4 & 2.2 & $t_7$ \\
        7 & 3.4 & 2.2 & $t_8$ \\
        7 & 3.4 & 2.7 & $t_9$ \\
        \hline
    \end{tabular}
    \label{tbl:decoupledliftingpath}
\end{table}
        
\begin{table}[tbp]
    \centering
    \caption{Tower crane lifting operations of the lifting path used in the trajectory planning simulation study.}
    \begin{tabular}{clc>{\centering\arraybackslash}m{0.2\columnwidth}}
        \hline
        \bf S. No. & \bf Operation & \bf Initial and Final Values & \bf Fixed Parameters \\
        \hline
        1 & 1\textsuperscript{st} Hoist & 5.5m ($d_{Hi}$) $\to$ 4.1m ($d_{Hf}$) & - \\
        2 & 1\textsuperscript{st} Trolley & 2.4m ($d_{Ti}$) $\to$ 2.6m ($d_{Tf}$) & $D_{H}$=4.1m \\
        3 & 1\textsuperscript{st} Slew & 83$^\circ$ ($\theta_{Si}$) $\to$ 66$^\circ$ ($\theta_{Sf}$) & $D_{H}$=4.1m, $D_{T}$=2.6m \\
        4 & 2\textsuperscript{nd} Hoist & 4.1m ($d_{Hi}$) $\to$ 3.3m ($d_{Hf}$) & - \\
        5 & 2\textsuperscript{nd} Trolley & 2.6m ($d_{Ti}$) $\to$ 3.3m ($d_{Tf}$) & $D_{H}$=3.3m \\
        6 & 3\textsuperscript{rd} Hoist & 3.3m ($d_{Hi}$) $\to$ 2.2m ($d_{Hf}$) & - \\
        7 & 3\textsuperscript{rd} Trolley & 3.3m ($d_{Ti}$) $\to$ 3.4m ($d_{Tf}$) & $D_{H}$=2.2m \\
        8 & 2\textsuperscript{nd} Slew & 66$^\circ$ ($\theta_{Si}$) $\to$ 7$^\circ$ ($\theta_{Sf}$) & $D_{H}$=2.2m, $D_{T}$=3.4m \\
        9 & 4\textsuperscript{th} Hoist & 2.2m ($d_{Hi}$) $\to$ 2.7m ($d_{Hf}$) & - \\
        \hline
    \end{tabular}
    \label{tbl:liftingoperations}
\end{table}

\subsubsection{Simulated trajectory results and analysis}
\label{subsec:LTPresults}

For all the 9 operations, the optimal values of operating time and normalized actuator effort satisfying all the constraints, obtained by the anti-swing trajectory planner are catalogued in Table \ref{tbl:optimalsLTP}. Utilizing the individual operating time, the time instants of the independent configurations of the tower crane (operation switching points), in its lifting path, can be deduced, as shown in Table \ref{tbl:timeinstantsLTP}. The total lifting time is computed as 74.88s (1min 14.88s).
 
\begin{table}[tbp]
    \centering
    \caption{Optimal solutions provided by the anti-swing trajectory planner for the tower crane operations in Table \ref{tbl:liftingoperations}.}
    \begin{tabular}{l>{\centering\arraybackslash}m{0.3\columnwidth}>{\centering\arraybackslash}m{0.3\columnwidth}}
        \hline
        \bf Operation & \bf Optimal Operating Time (s) & \bf Optimal Actuator Effort \\
        \hline
        1\textsuperscript{st} Hoist & 11.86 & 0.75 \\
        1\textsuperscript{st} Trolley & 4.36 & 0.76 \\
        1\textsuperscript{st} Slew & 5.36 & 0.45 \\
        2\textsuperscript{nd} Hoist & 6.77 & 1.31 \\
        2\textsuperscript{nd} Trolley & 7.14 & 0.67 \\
        3\textsuperscript{rd} Hoist & 9.33 & 0.95 \\
        3\textsuperscript{rd} Trolley & 3.25 & 0.42 \\
        2\textsuperscript{nd} Slew & 22.16 & 1.01 \\
        4\textsuperscript{th} Hoist & 4.65 & 1.59 \\
        \hline
    \end{tabular}
    \label{tbl:optimalsLTP}
\end{table}

\begin{table}[tbp]
    \centering
    \caption{Time instants of lifting configurations according to the optimal solutions provided by the anti-swing trajectory planner for the tower crane lifting path in Table \ref{tbl:decoupledliftingpath}.}
    \begin{tabular}{cc}
        \hline
        \bf Time & \bf Value \\
        \hline
        \bf $t_0$ & 0s \\
        \bf $t_1$ & 11.86s \\
        \bf $t_2$ & 16.22s \\
        \bf $t_3$ & 21.58s \\
        \bf $t_4$ & 28.35s \\
        \bf $t_5$ & 35.49s \\
        \bf $t_6$ & 44,82s \\
        \bf $t_7$ & 48.07s \\
        \bf $t_8$ & 70.23s \\
        \bf $t_9$ & 74.88s \\
        \hline
    \end{tabular}
    \label{tbl:timeinstantsLTP}
\end{table}

\begin{figure}[tbp]
    \begin{center}
        \includegraphics[width=\columnwidth]{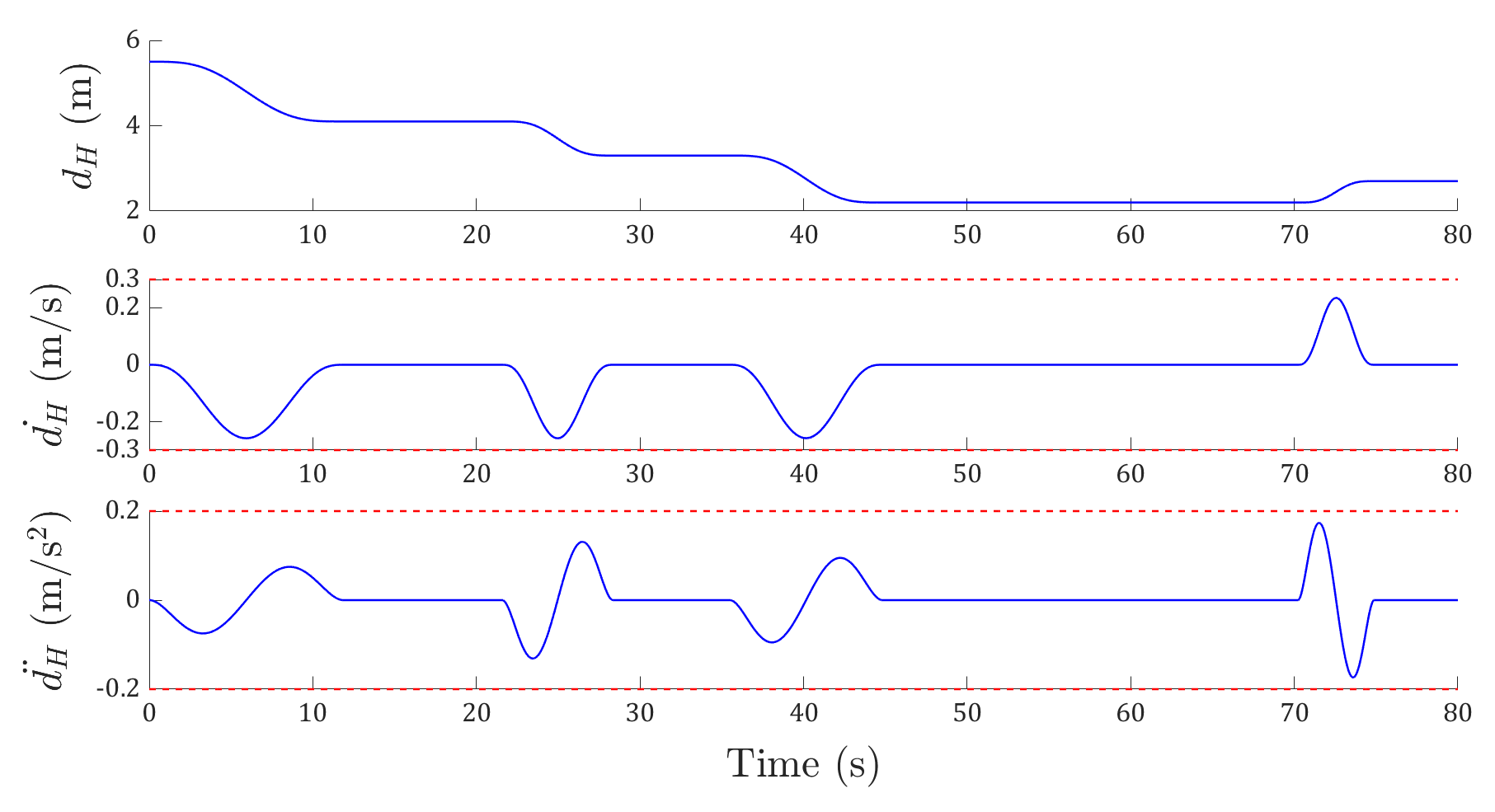}
        \caption{Hoist operation trajectories obtained from the trajectory planner for the tower crane lifting path used in the trajectory planning simulation study. (Red dashed lines indicate the mechanical constraint limits on the hoisting velocity and acceleration.)}
        \label{fig:hoisttrajectories}
    \end{center}
\end{figure}

\begin{figure}[tbp]
    \begin{center}
        \includegraphics[width=\columnwidth]{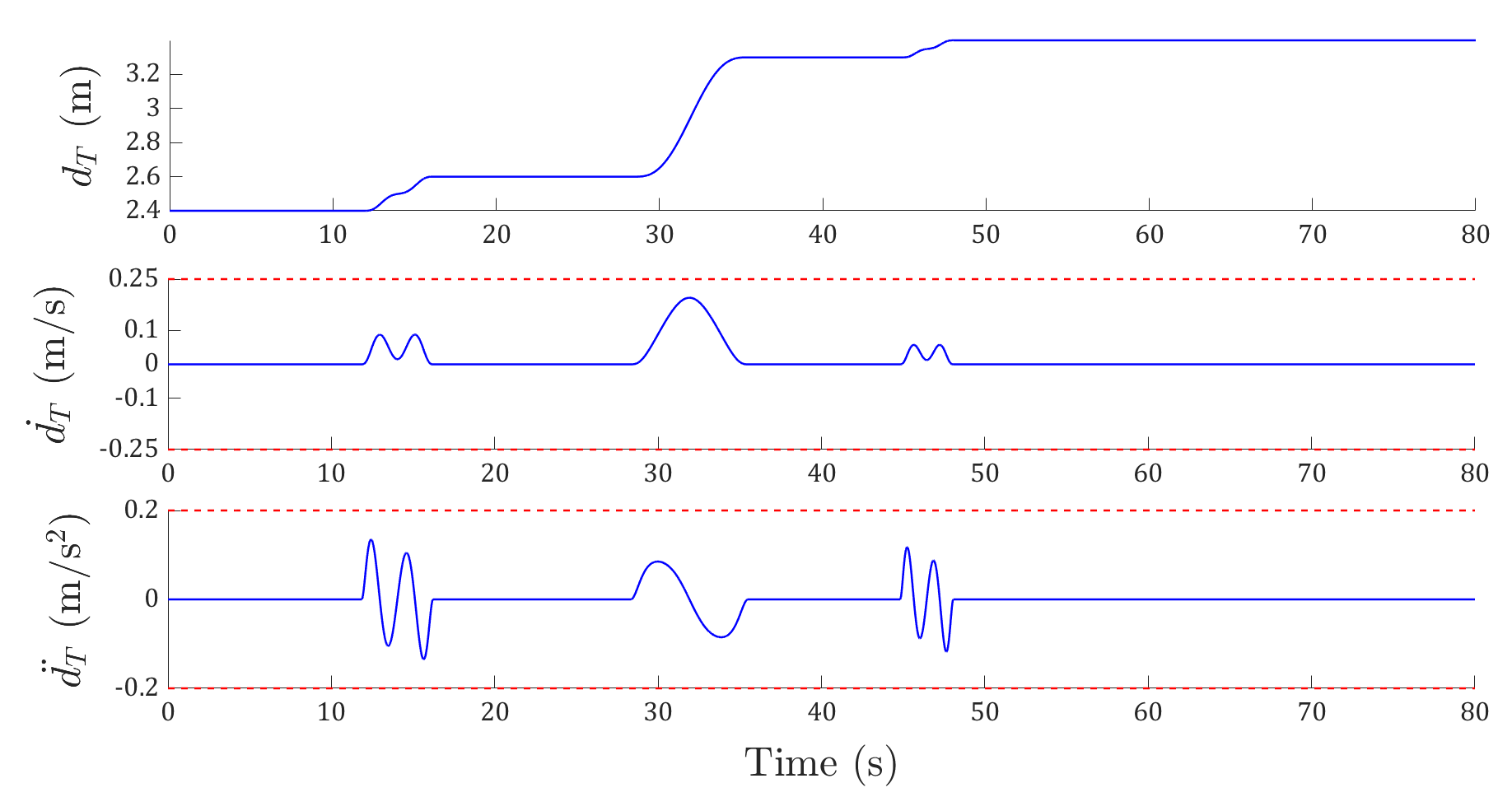}
        \caption{Trolley operation trajectories obtained from the trajectory planner for the tower crane lifting path used in the trajectory planning simulation study. (Red dashed lines indicate the mechanical constraint limits on the trolley velocity and acceleration.)}
        \label{fig:trolleytrajectories}
    \end{center}
\end{figure}

\begin{figure}[tbp]
    \begin{center}
        \includegraphics[width=\columnwidth]{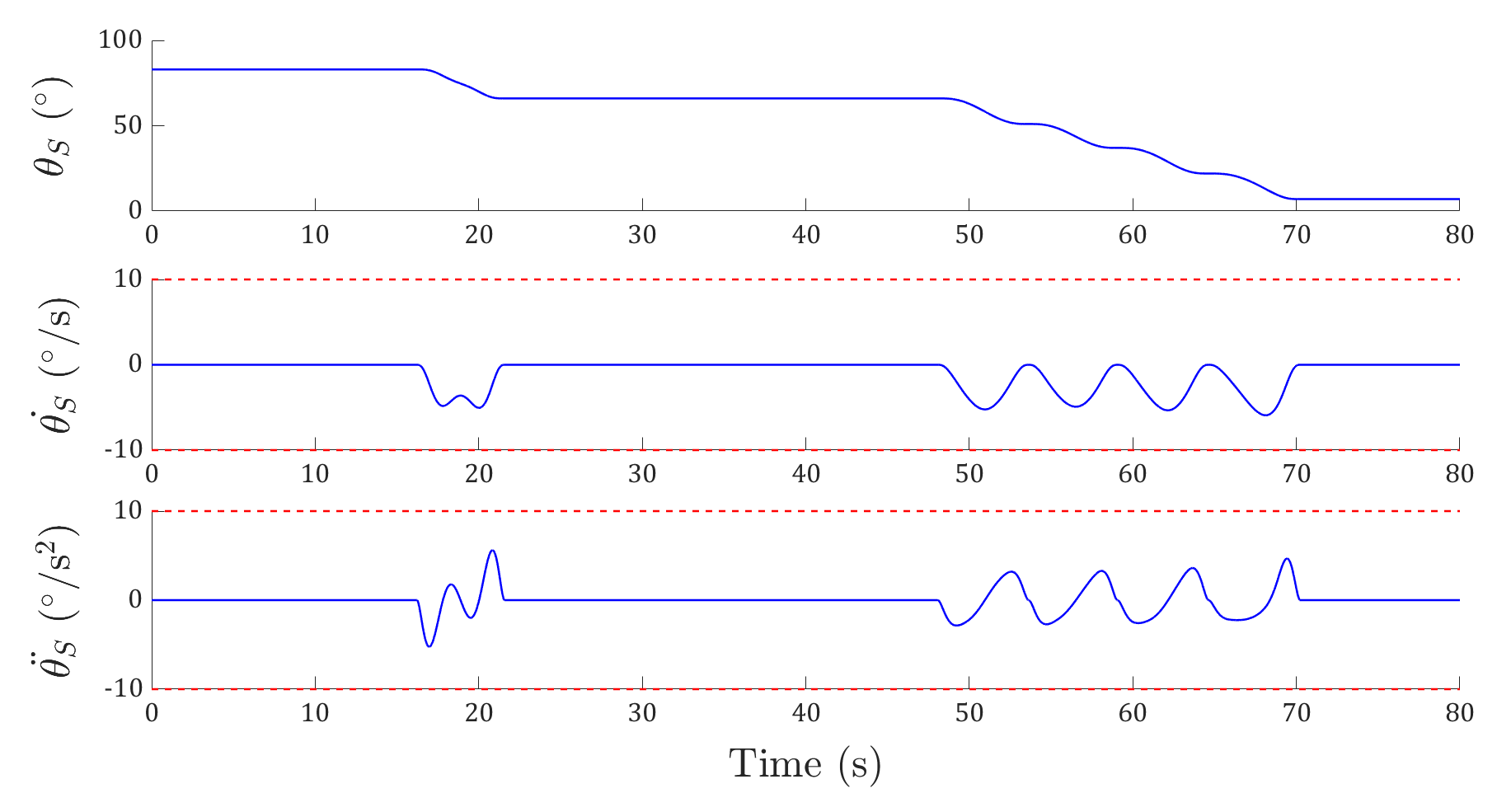}
        \caption{Slew operation trajectories obtained from the trajectory planner for the tower crane lifting path used in the trajectory planning simulation study. (Red dashed lines indicate the mechanical constraint limits on the slew velocity and acceleration.)}
        \label{fig:slewtrajectories}
    \end{center}
\end{figure}

The optimal hoist operation trajectories throughout the lifting path are plotted in Figure \ref{fig:hoisttrajectories}. The velocity and acceleration of the hoist motion are kept well within their limits for the whole lifting path. Trajectories for the trolley operations are presented in Figure \ref{fig:trolleytrajectories}. Optimal trolley velocity and acceleration are also bound by the mechanical limits, as specified. The trolley velocity is always unidirectional throughout an operation (positive for all the 3 cases since all trolley operations are away from the tower along the jib), indicating that the trolley never changes direction during actuation. For the slew operation, the optimal trajectories of the jib are provided in Figure \ref{fig:slewtrajectories}, where the slew motion can be seen to abide by the constraints of the tower crane slew actuator. Slew velocity is always negative as both the slew operations are performed in the clockwise direction about the $z$ axis (from the perspective of the crane top view). Nevertheless, the sign of the slew velocity never changes, implying that the motions are unidirectional.

\begin{figure}[tbp]
    \begin{center}
        \includegraphics[width=\columnwidth]{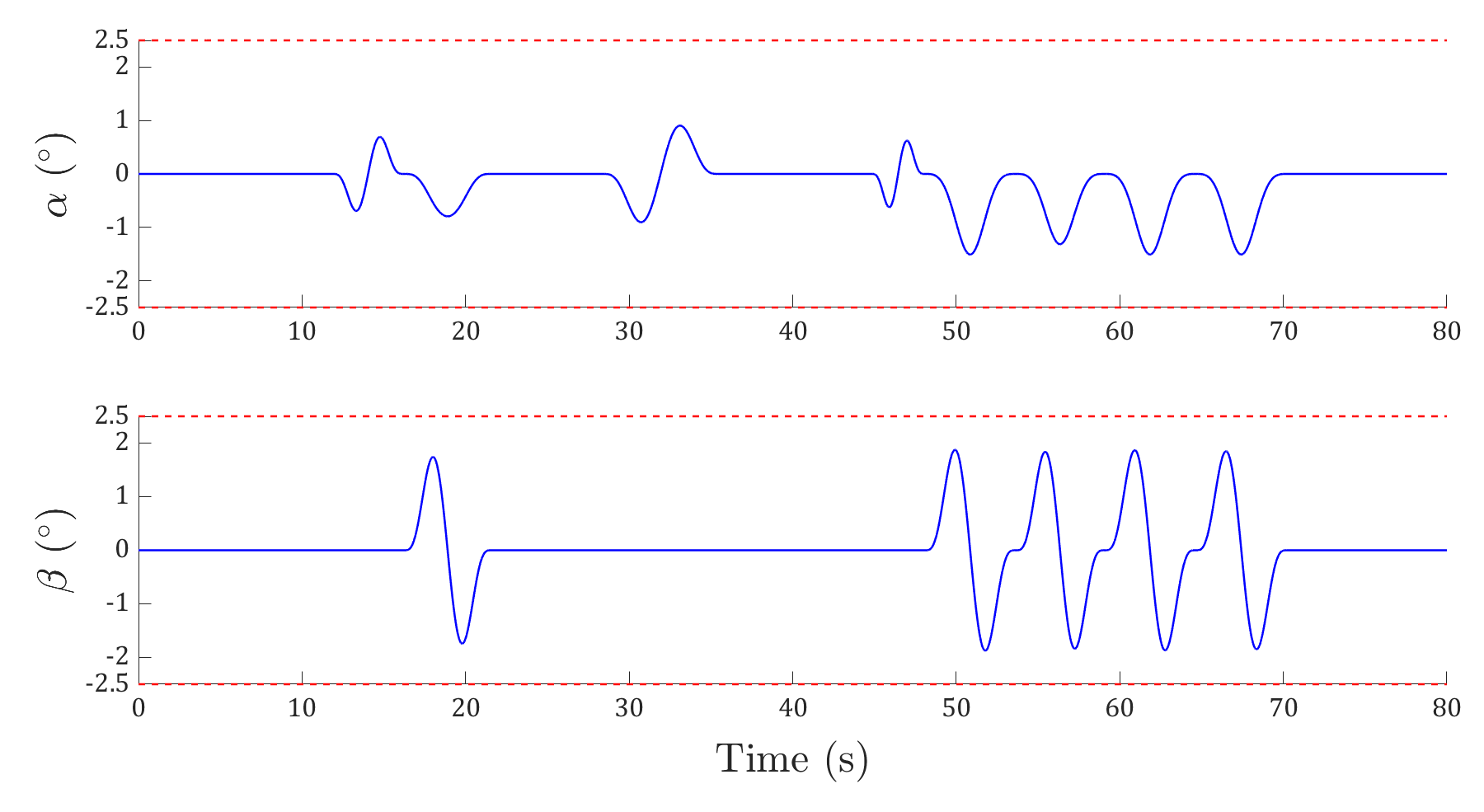}
        \caption{Radial ($\alpha$) and tangential ($\beta$) payload swing trajectories obtained from the trajectory planner for the tower crane lifting path used in the trajectory planning simulation study. (Red dashed lines indicate the safety constraint limits on the payload swing angles.)}
        \label{fig:swingtrajectories}
    \end{center}
\end{figure}

The radial and tangential payload swing angles for the entire lifting path are shown in Figure \ref{fig:swingtrajectories}. As evident, both the swing angles are kept well within the safe ranges specified by the permissible payload deflection to realize the optimal collision-free planned lifting path. Contiguous operations have no radial payload swing $\alpha$ at the operation switching points in between. Also, there is no residual swing after the crane operations. It is to be noted, that these are offline optimal reference trajectories, hence do not address any external disturbances (such as strong wind during lifting) or parametric uncertainties. To tackle such situations, a feedback tracking controller is needed to compensate for the output due to the disturbance inputs. Nevertheless, the proposed anti-swing trajectory planner has been shown as an effective and efficient open-loop control component in planning multi-objective optimal reference trajectories for tower cranes, while constraining payload swing during lifting and eliminating residual swing. Such optimal reference trajectories can be used by the tower crane controller to track the planned optimal collision-free lifting path.

\section{Conclusion}
\label{sec:conclusion}

Being under-actuated systems, tower cranes produce unactuated payload motion during lifting, which needs to be reduced to ensure optimal and safe execution of lifting paths in autonomous construction environments. For the trolley motion, the payload behaves as a planar pendulum, while during slew operation, it exhibits spherical pendulum behavior. The present research work proposes an anti-swing tower crane trajectory planner, which plans time-energy optimal lifting trajectories while addressing all the transient constraints imposed due to mechanical and safety considerations. A robotized model of the tower crane is prepared to analyze the dynamics of the crane lifting motions. Elaborate analysis of the equations of motion establishes the system as highly non-linear with coupling between the actuated and unactuated DOFs. This coupling behavior is dealt with the utilization of carefully considered auxiliary DOFs, which are the payload positions during lifting. The differential flatness of the tower crane system is proved for the trolley and slew motions, which generate the additional DOFs. 7\textsuperscript{th} and 11\textsuperscript{th} degree B\'{e}zier curves are implemented to interpolate the parameterized trajectories of the hoisting height, and the flat outputs for the trolley and slew operations, respectively. Based on this, the trajectory planning tasks for the fundamental crane operations are formulated as constrained multi-objective optimization problems. The actuated motions are limited according to the mechanical constraints of the crane, and the payload swing amplitudes are restricted within the safe interval of payload deflection. Elimination of the residual swing is considered via the simultaneous constraints of zero position and velocity of the radial and tangential swing angles.

Two MOEAs, namely NSGA-II and GDE3, are assessed to select the optimizer that can provide a wide range of Pareto optimal solutions in near real-time. All three fundamental crane operations are taken into consideration while comparing the performances of the MOEAs. GDE3 emerges as the preferred solver in this case. After solving the MOTOPs, the crane operation trajectories are then computed via the corresponding planned flat output trajectories. Simulation studies emulating real-world lifting scenarios are conducted to verify the effectiveness of the proposed strategy. The designed anti-swing trajectory planner is incorporated in a CALP system built for autonomous crane lifting in complex construction environments. Trajectories for all the operations of a complete collision-free lifting path of the tower crane are calculated via the developed anti-swing trajectory planner. The simulated results demonstrate that the planner can produce optimal jerk-continuous trajectories keeping all the state variables within their respective limits and reducing the residual payload swing completely. However, as an offline trajectory planning method, the current approach is unable to address the parametric uncertainties of the system as well as the effect of external disturbances on the payload motions. To address these issues, development of a feedback tracking controller is aimed as a future work, which can take the optimal trajectories generated by the proposed planner as input and execute the lifting path accurately, eliminating the effects of the aforementioned scenarios.

\bibliographystyle{IEEEtran}
\bibliography{preprint}

\end{document}